\documentclass[10pt,journal,compsoc]{IEEEtran}

\ifCLASSOPTIONcompsoc
  \usepackage[nocompress]{cite}
\else
  \usepackage{cite}
\fi

\usepackage[utf8]{inputenc}
\usepackage[T1]{fontenc}
\usepackage{hyperref}
\usepackage{url}
\usepackage{nicefrac}
\usepackage{microtype}
\usepackage{xcolor}
\usepackage{enumitem}
\usepackage{multirow}
\usepackage{longtable}
\usepackage{caption}
\usepackage{amsmath,amsfonts,amssymb,amsthm}
\usepackage{graphicx}
\usepackage{booktabs}
\usepackage[ruled,vlined,linesnumbered]{algorithm2e}
\usepackage{siunitx,threeparttable}
\usepackage{subcaption}
\usepackage{tabularx}
\usepackage{rotating}

\begin{document}
\title{Understanding and Stabilizing Deep Q-Learning via Controlled Bootstrapping and Regulated Value Dynamics}
\author{Bozhou~Chen, Yongyi~Wang, Hanyu~Liu, Xionghui~Yang, and Wenxin~Li%
\IEEEcompsocitemizethanks{\IEEEcompsocthanksitem
All authors are with the School of Computer Science, Peking University, Beijing, China.\protect\\
Corresponding author: Wenxin Li.\protect\\
E-mail: 2301111899@stu.pku.edu.cn; wangyongyi@pku.edu.cn; 2301111944@stu.pku.edu.cn; yangxionghui@stu.pku.edu.cn; lwx@pku.edu.cn}}

\IEEEtitleabstractindextext{%
\begin{abstract}
Deep Q-learning (DQL) has achieved remarkable empirical success in reinforcement learning, yet its training process remains notoriously unstable. Existing studies often attribute instability to isolated factors such as overestimation bias or representation learning issues, lacking a unified understanding of how different sources of instability interact during recursive value estimation. In this work, we provide a systematic analysis of instability in deep Q-learning from three complementary perspectives: operator-level bias in Bellman bootstrapping, estimator-level sensitivity of greedy action selection to regression noise, and parameter-dynamics imbalance under aggressive data reuse. We identify a reward-triggered self-reinforcing trap and characteristic parameter spike dynamics, then derive stabilization principles for controlled bootstrapping, ensemble quantile estimation, and spike-based parameter regulation. Experiments on Atari-100K and Procgen demonstrate competitive performance and improved training stability.
\end{abstract}
}

\maketitle
\IEEEdisplaynontitleabstractindextext
\IEEEpeerreviewmaketitle

\IEEEraisesectionheading{\section{Introduction}\label{sec:introduction}}
\IEEEPARstart{D}{eep} Q-learning has become a cornerstone method in deep reinforcement learning (RL), widely applied in domains ranging from video games to robotics. Its appeal lies in learning effective policies directly from high-dimensional observations via neural network approximation.
% with widespread applications in domains ranging from video games to robotics. Its appeal lies in its ability to learn effective policies directly from high-dimensional observations through neural network approximation. Despite its empirical success, training deep RL agents remains notoriously unstable. Value estimates may fluctuate dramatically, grow uncontrollably, or plateau prematurely, often making learning dynamics difficult to interpret or control.

A large body of prior work has studied instability in value learning from specific perspectives. Classical analyses focus on overestimation bias induced by the max operator in Bellman updates. Standard Q-learning applies the maximization operator to noisy value estimates, which leads to systematically optimistic targets. Double Q-learning~\cite{hasselt2010double} and its deep variant Double DQN~\cite{van2016deep} mitigate this issue by decoupling action selection from evaluation. More recent methods further reduce bias or variance through ensemble estimation or distributional value modeling.

However, empirical evidence suggests that instability in deep Q-learning cannot be fully explained by maximization bias alone. In practice, value learning exhibits complex feedback effects involving bootstrap targets, stochastic regression, and evolving network representations. These mechanisms interact with each other through the recursive structure of value updates, yet they are typically studied in isolation.

In this work, we revisit instability in deep Q-learning from a unified perspective and identify three interacting mechanisms that jointly shape learning dynamics:

\textbf{Operator-level instability.}
Recursive Bellman bootstrapping may introduce structural bias in target construction. Beyond classical overestimation, we observe a previously overlooked feedback mechanism, which we term the self-reinforcing trap. In reward-bearing transitions, value amplification induced by positive rewards interacts with action-conditioned representation generalization. As a result, the same action may be repeatedly selected in bootstrap targets, creating a feedback loop that continuously amplifies its value.

\textbf{Estimator-level instability.}
Deep Q-learning is fundamentally a regression problem in which the value function is learned through stochastic approximation. Even when estimation noise is unbiased, greedy action selection depends on relative value differences. When action gaps are small, regression noise may alter greedy decisions, which subsequently changes the distribution of collected experience. In this way, estimation noise propagates through the control loop and may destabilize learning dynamics.

\textbf{Parameter-dynamics instability.}
We further observe that neural networks used for value approximation may gradually develop imbalanced parameter distributions during training, particularly under high replay ratios. In such regimes, a small subset of parameters may grow disproportionately large while the majority remain relatively unchanged. This phenomenon is closely related to the loss of network plasticity~\cite{nikishin2022primacy, sokar2023dormant, lyle2023understanding}. To better characterize this effect, we introduce a simple diagnostic statistic termed the spike ratio, which measures the concentration of extreme parameter magnitudes within each layer and serves as a practical indicator of parameter imbalance.

% These observations suggest that instability in deep Q-learning arises from interacting feedback mechanisms across target construction, value estimation, and parameter evolution. Stabilizing value learning therefore requires coordinated control over these components rather than isolated heuristics.
These observations suggest that instability in deep Q-learning arises from interacting feedback mechanisms across target construction, value estimation, and parameter evolution, requiring coordinated control rather than isolated heuristics.
Based on this analysis, we derive several practical stabilization principles for value learning. These include controlling recursive amplification in bootstrap targets, reducing decision variance in value regression, and maintaining parameter plasticity during prolonged training. We instantiate these principles in a practical learning framework that integrates controlled bootstrapping, ensemble quantile regression, and parameter regulation.

Extensive experiments on Atari-100K and Procgen demonstrate that the resulting algorithm achieves strong performance and improved training robustness. More importantly, the empirical results support the proposed instability analysis and illustrate how the identified mechanisms influence learning dynamics in practice.

Our main contributions are summarized as follows:
\begin{itemize}
    \item We provide a systematic analysis of instability in deep Q-learning, identifying three interacting mechanisms: operator-level bias in bootstrapping, estimator-level decision sensitivity to regression noise, and parameter-dynamics imbalance during training.
    \item We uncover a previously overlooked feedback mechanism termed the self-reinforcing trap, where reward-driven value amplification interacts with representation generalization to produce biased bootstrap dynamics.
    \item We reveal the role of parameter spike dynamics in the loss of network plasticity and introduce the spike ratio as a practical statistic for monitoring training stability.
    \item Based on these insights, we derive stabilization principles for value learning and instantiate them in a practical algorithm that achieves competitive performance on Atari-100K and Procgen benchmarks.
\end{itemize}

The remainder of this paper is as follows. Section~\ref{sec: rel} reviews related work on bootstrapping bias, distributional learning, ensemble estimation, etc. Section~\ref{sec:analyse} presents the instability analysis from operator-level, estimator-level, and parameter-dynamics perspectives. Section~\ref{sec:method} introduces stabilization principles derived from this analysis, and Section~\ref{sec: algo} presents their algorithmic instantiation.
Section~\ref{sec: exp} provides experimental evaluation and mechanism studies. Finally, Section~\ref{sec: con} concludes the paper and discusses future directions.
\section{Related Work\label{sec: rel}}
\subsection{Overestimation and Stable Bootstrapping Targets}

The maximization operator in the Bellman update introduces systematic overestimation when applied to noisy value predictions.
Thrun and Schwartz~\cite{thrun1993issues} first identified this bias in the tabular setting, and under function approximation such bias may accumulate through recursive bootstrapping.
Double Q-learning~\cite{hasselt2010double} mitigates this issue by decoupling action selection from evaluation using two estimators, and Double DQN~\cite{van2016deep} extends this idea to deep RL using online and target networks.
Subsequent works further refine bootstrap target construction.
Averaged-DQN~\cite{anschel2017averaged} reduces variance by averaging past networks, while Maxmin DQN~\cite{lan2020maxmin} selects the minimum among an ensemble of critics to control optimism.
In continuous control, TD3~\cite{fujimoto2018addressing} adopts clipped double Q-learning, taking the minimum of two critics as a conservative target.

Ensemble-based strategies provide additional bias control.
REM~\cite{agarwal2020optimistic} enforces Bellman consistency on random convex combinations of Q-heads, and REDQ~\cite{chen2021redq} leverages larger ensembles with randomized subsampling to stabilize training under high update-to-data ratios.

These approaches primarily stabilize value learning at the level of bootstrap target construction by reducing maximization bias or target variance. However, they mainly address the bias properties of the Bellman target itself, while other factors affecting the dynamics of recursive value learning remain less explored.

\subsection{Distributional Reinforcement Learning and Ensemble Estimation}

Distributional RL models, such as C51~\cite{bellemare2017distributional}, the full return distribution rather than only its expectation, capturing richer information about uncertainty in value estimation.
C51 introduced a categorical representation over a fixed support and demonstrated improved stability on Atari.
QR-DQN~\cite{dabney2017distributional} replaced categorical projections with quantile regression, while IQN~\cite{dabney2018implicit} parameterized the entire quantile function to allow flexible distributional modeling.
In continuous control, TQC~\cite{kuznetsov2020controlling} combines distributional critics with an ensemble of Q-functions and truncates the highest quantile atoms to suppress overestimation, providing a softer alternative to scalar minimum operators.

Ensemble techniques improve both value stability and exploration.
Bootstrapped DQN~\cite{osband2016deep} trained multiple Q-heads with bootstrapped data partitions, enabling temporally consistent deep exploration.
REM~\cite{agarwal2020optimistic} performs ensemble aggregation across critics to obtain more accurate value estimates.
SUNRISE~\cite{lee2021sunrise} introduced uncertainty-weighted Bellman backups, down-weighting unreliable targets to reduce error propagation.
SPQR~\cite{lee2023spqr} explicitly addressed ensemble collapse by injecting structured noise to preserve independent estimation behavior.

These methods improve robustness at the estimation level by modeling return distributions or aggregating multiple value predictors.
However, they are primarily motivated by improving prediction accuracy or uncertainty estimation, while the role of regression noise in affecting decision and interaction dynamics remains less explicitly analyzed.

\subsection{Value Learning in Data-Efficient Regimes}

Recent progress in data-efficient reinforcement learning has significantly improved performance under limited interaction budgets, such as the Atari-100K setting.
A central trend in this regime is the integration of strong representation learning objectives into value-based methods.

CURL~\cite{laskin2020curl} introduced contrastive representation learning for pixel-based RL, improving sample efficiency through latent consistency.
SPR~\cite{schwarzer2020data} and SR-SPR~\cite{d2022sample} further leveraged self-predictive representations by enforcing temporal consistency in latent space.
DrQ~\cite{yarats2021image} demonstrated that simple data augmentation substantially improves data efficiency and generalization.
More recent methods such as BBF~\cite{schwarzer2023bigger}, SGF~\cite{robine2025simple}, Drama~\cite{wang2024drama}, IRIS~\cite{micheli2022transformers}, and STORM~\cite{zhang2023storm} combined stronger regularization, auxiliary objectives, and architectural refinements to further push data-efficient performance.

In addition to representation learning, high replay ratios and aggressive update-to-data strategies have been widely adopted in data-limited regimes.
DER~\cite{van2019use} and BBF~\cite{schwarzer2023bigger} showed that increasing gradient updates per environment interaction can substantially improve sample efficiency.
However, heavy data reuse may also amplify bootstrap errors and increase sensitivity to estimation noise, particularly under recursive value updates.

% These methods substantially improve sample efficiency through enhanced representation learning and increased data reuse.
% However, these factors are not the essence of instability, which lies in unstable bootstrapping, even in the data-efficient regimes.
% In this paper, we show that competitive performance can be achieved even without relying on these data-efficient techniques.

These methods substantially improve sample efficiency through enhanced representation learning and increased data reuse. However, instability still fundamentally originates from recursive bootstrapping dynamics.

\subsection{Optimization Dynamics and Network Plasticity}

Beyond bias in bootstrap targets, a growing body of work has investigated instability in deep reinforcement learning from the perspective of optimization dynamics and network behavior during training.
Early stabilization techniques primarily aim to mitigate the feedback effects introduced by bootstrapped updates.  
Target networks~\cite{mnih2015human} provide a slowly updated objective to reduce oscillations in value estimates.  
Additional techniques such as the Huber loss~\cite{mnih2015human} limit the influence of extreme TD errors, and safe backup operators including Retrace($\lambda$) truncate or down-weight multi-step returns to prevent uncontrolled value propagation.

Subsequent studies examined instability from the perspective of optimization under function approximation.
Spectral normalization~\cite{gogianu2021spectral} constrains the Lipschitz constant of Q-networks to stabilize value prediction.
Creus~Castanyer et al.~\cite{castanyer2025stable} identified unstable gradient flow in deep Q-networks under non-stationary data and proposed architectural and optimization modifications.
Related analyses~\cite{markowitz2024avoiding, nauman2024overestimation} showed that off-policy algorithms remain sensitive to estimation errors amplified through bootstrapped updates.

Another line of work focuses on the loss of network plasticity during prolonged reinforcement learning training.  
Nikishin et al.~\cite{nikishin2022primacy} identified the primacy bias, where agents overfit early interaction data and struggle to adapt under high replay ratios.
Sokar et al.~\cite{sokar2023dormant} reported the dormant neuron phenomenon, observing that a large fraction of neurons become inactive during training.
Lyle et al.~\cite{lyle2023understanding} further connected this loss of plasticity to rank collapse in learned representations, demonstrating that common architectural and optimization choices can accelerate representational degradation.

To mitigate plasticity loss, several intervention strategies have been proposed.
Periodic parameter resets~\cite{nikishin2022primacy} aim to counteract primacy bias but may discard useful structure.
Shrink-and-Perturb~\cite{ash2020warm} restores gradient flow by shrinking parameters toward zero while injecting noise, and ReDo~\cite{sokar2023dormant} selectively reinitializes dormant neurons based on activation statistics to recycle unused network capacity.

Together, these studies highlight that instability in deep reinforcement learning is closely related to optimization dynamics and the evolution of network representations during prolonged bootstrapped training.

\subsection{World Models and Model-Based Reinforcement Learning}

Recent years have seen rapid advances in model-based reinforcement learning, particularly in data-efficient settings such as Atari-100K.
These methods improve sample efficiency by learning an explicit environment model and performing planning in latent space.
IRIS~\cite{micheli2022transformers} introduced a Transformer-based world model operating on discrete VQ-VAE latents, achieving strong sample efficiency without explicit lookahead search.
DreamerV3~\cite{hafner2023mastering} scaled recurrent latent world models into a unified algorithm spanning over 150 tasks, emphasizing normalization and architectural stability to ensure cross-domain robustness.
STORM~\cite{zhang2023storm} combined Transformer architectures with stochastic latent variables to enhance modeling flexibility, while DIAMOND~\cite{alonso2024diffusion} leveraged diffusion-based generation to better preserve visual details that discrete latent models may discard.
DART~\cite{agarwal2024learning} and Drama~\cite{wang2024drama} further explored trajectory-aware and state-space modeling architectures, continuing the trend toward increasingly expressive and stable world models.
% Unlike these approaches, which rely on learning an explicit dynamics model to guide value estimation or planning, our work remains strictly model-free and focuses on stabilizing recursive value learning itself.
These methods address sample efficiency through model learning,
while our work focuses on the stability of value learning itself.
\section{Analysis of Instability Mechanisms\label{sec:analyse}}
% In this section, we analyze value instability from three complementary perspectives. First, at the operator level, structural properties of the Bellman backup introduces systematic bias in target construction. Second, at the estimator level, even unbiased value regression can lead to unstable action selection due to decision sensitivity to estimation noise. Third, from the parameter-dynamics perspective, optimization under non-stationary sampling can distort parameter distributions and reduce representational plasticity. Together, these mechanisms characterize the primary sources of instability in deep Q-learning.
Although instability in deep Q-learning has long been recognized, existing explanations typically focus on individual phenomena such as maximization bias or representation learning effects. In practice, however, value learning forms a recursive feedback system in which multiple components interact through the Bellman update and environment interaction loop. Instability therefore often arises not from a single source, but from the coupling of several mechanisms that influence the dynamics of recursive value estimation.

In this section, we analyze instability in deep Q-learning from three complementary perspectives corresponding to different stages of the value learning process. First, at the operator level, recursive Bellman bootstrapping may introduce structural bias in target construction, leading to feedback amplification effects beyond classical maximization bias. Second, at the estimator level, the stochastic nature of value regression may perturb greedy action selection, causing regression noise to propagate through the control loop and alter the data distribution encountered during training. Third, at the parameter-dynamics level, neural network parameters may gradually evolve toward imbalanced configurations under aggressive data reuse, reducing the network's ability to adapt to changing value targets.

These mechanisms interact through the recursive structure of value learning. Amplified bootstrap targets influence value regression, noisy estimation alters control decisions and experience collection, and evolving parameter configurations further shape the learning dynamics. Together, these feedback pathways create complex instability patterns that are difficult to explain from any single perspective. The following subsections examine these mechanisms in detail and provide empirical evidence illustrating their roles in deep Q-learning instability.

\subsection{Operator-Level Instability: Unstable Recursive Bootstrapping\label{sec: operator_level}}
Operator-level instability arises from structural biases in the Bellman backup.
Standard Q-learning constructs targets via
\begin{equation}
r + \gamma \max_{a'} Q_{\bar{\theta}}(s',a').
\end{equation}
Because the maximum is taken over noisy estimates,
\begin{equation}
\label{equ: decouple}
\mathbb{E}\!\left[\max_{a} \hat{Q}(s,a)\right]
\;\ge\;
\max_{a} \mathbb{E}\!\left[\hat{Q}(s,a)\right]
\end{equation}
leading to classical overestimation bias.

In deep reinforcement learning, an additional source of overestimation arises from function approximation and representation generalization.
In discrete control, the action-value function is commonly parameterized as
\begin{equation}
Q_\theta(s,a) = f_{\theta_1}\!\left(\phi_{\theta_2}(s), a\right),
\end{equation}
where $\phi_{\theta_2}(\cdot)$ denotes a shared state encoder and $f_{\theta_1}(\cdot,a)$ is an action-conditioned output head.
Under this architecture, generalization primarily occurs across states for the same action, while value estimates for different actions are modeled by separate heads and are therefore not directly coupled by representation generalization.

As a result, for a fixed action $a$, the corresponding value estimates satisfy $Q_\theta(s,a) \approx Q_\theta(s',a)$, whereas no such coupling is induced between $Q_\theta(s,a)$ and $Q_\theta(s',a')$ for $a'\neq a$.
This action-conditioned generalization improves sample efficiency, but it also creates an implicit pathway through which bootstrapped targets can propagate across nearby states for the same action.

A particularly problematic case arises for reward-bearing transitions.
Consider a replayed transition $(s,a,r,s')$ with $r>0$.
Due to action-conditioned generalization, the update of $Q(s,a)$ driven by the positive reward also affects the estimate of $Q(s',a)$, since the two states induce similar representations and share the same action head.
As a result, after a number of updates, the value $Q(s',a)$ tends to become relatively large compared to other actions at state $s'$.

\begin{figure}[!t]
    \centering
    \includegraphics[width=\linewidth]{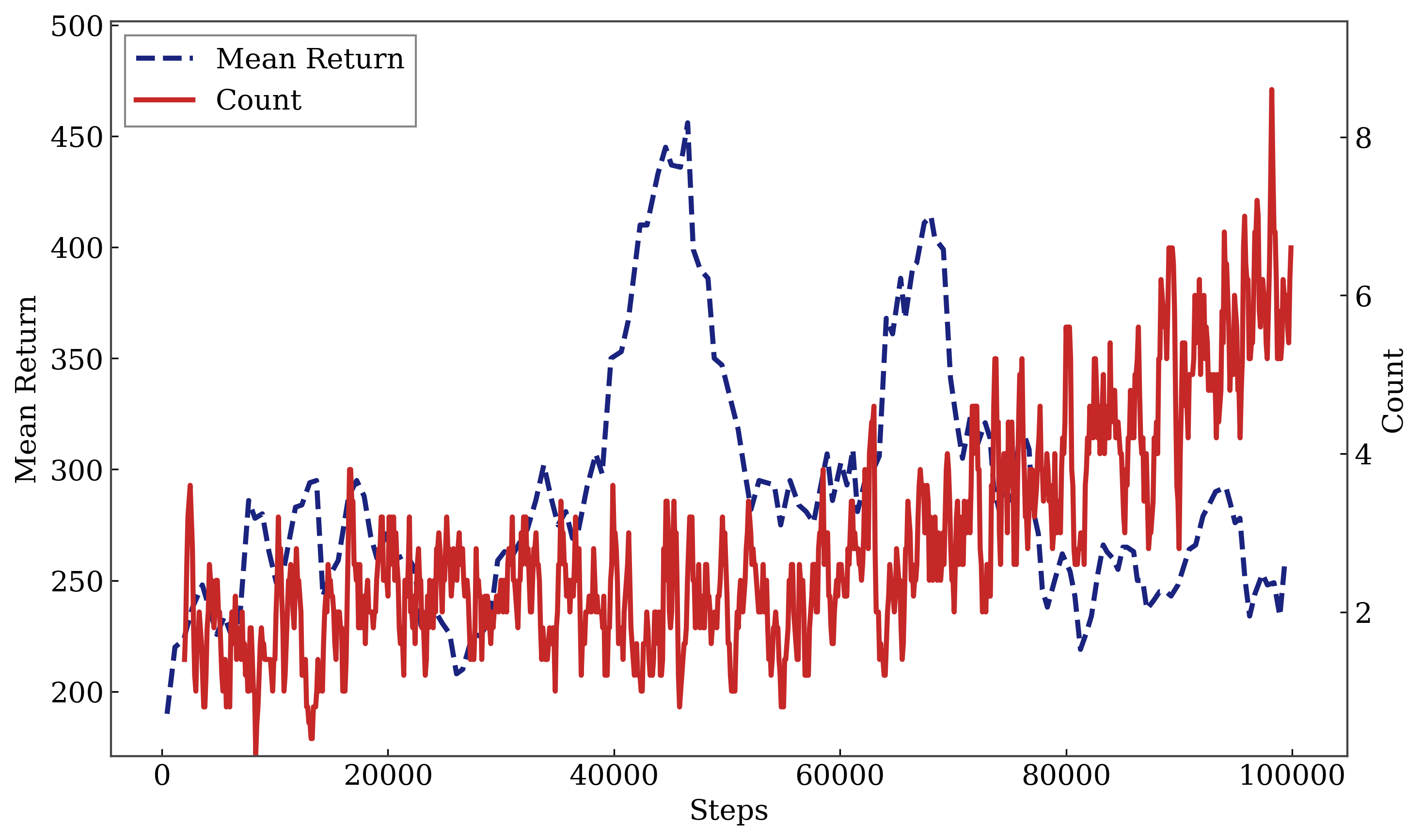}
    \caption{Reward-triggered self-reinforcing bias in Atari \textit{Alien}.
    The dashed blue curve (left axis) shows the recent 10-episode mean return during training, while the solid red curve (right axis) reports the minibatch statistic $\mathrm{Count}\!\left(a = \arg\max_{a'} Q(s', a')\right)$, the number of sampled transitions whose greedy bootstrap action at the next state coincides with the original action. 
    The action space contains 18 actions and the minibatch size is 32. 
    The consistently elevated counts indicate a strong coupling between $a$ and the bootstrap action $a'^*$, providing empirical evidence for the self-reinforcing trap discussed in this section.}
    \label{fig: nexta}
\end{figure}
When computing the bootstrapped target, the max operator selects the action $a'^* \in \arg\max_{a'} Q(s',a')$.
Because $Q(s',a)$ has already been amplified through representation generalization induced by the positive reward $r>0$, the maximization step is biased toward selecting the same action, i.e., $a'^* = a$.
This bias is not caused by the max operator alone, but the interaction between reward-driven value amplification and action-conditioned generalization.
As illustrated in Fig.~\ref{fig: nexta}, we empirically examine this phenomenon in the Atari \textit{Alien} environment by recording the number of transitions for which $a = \arg\max_{a'} Q(s', a')$.
The results show a pronounced bias toward selecting the previous action, indicating a strong correlation between $a'^*$ and $a$.

Once $a'^* = a$ is selected, the resulting target $r + \gamma Q(s',a)$ further increases $Q(s,a)$.
This creates a positive feedback loop in which the reward-induced increase of $Q(s,a)$ propagates to $Q(s',a)$, and the maximization step repeatedly selects the same action.
Importantly, this mechanism is specific to transitions with $r>0$.
For transitions with zero reward, no persistent upward shift is introduced, and the max operator does not exhibit the same systematic preference for reselecting the original action.

Notably, the magnitude of this self-reinforcing amplification admits an upper bound in an idealized repeated-replay scenario.
Consider repeatedly replaying the same reward-bearing transition $(s,a,r,s')$ with $r>0$, and suppose the maximization step persistently reselects the same action so that $a'^*=a$ holds throughout training, as indicated by our empirical observation.
In this case, the bootstrapped update reduces to a one-dimensional fixed-point iteration
\begin{equation}
Q_{k+1}(s,a) \;=\; r + \gamma\, Q_k(s',a).
\end{equation}
Under the self-reinforcing coupling described above, $Q_k(s',a)$ is repeatedly driven upward in tandem with $Q_k(s,a)$, and the iteration effectively behaves as
\begin{equation}
Q_{k+1}(s,a) \;\approx\; r + \gamma\, Q_k(s,a),
\end{equation}
whose unique fixed point is
\begin{equation}
Q(s,a) \;\to\; \frac{r}{1-\gamma}.
\end{equation}
Therefore, even though the feedback loop can systematically inflate $Q$-values, its amplification under repeated replay of a single transition is bounded and converges to a finite limit determined by the reward scale and discount factor.

\subsection{Estimator-Level Instability: Interaction Drift Induced by Regression Noise\label{sec: estimator_level}}
Even when the Bellman operator is unbiased and target construction is structurally stable, 
deep Q-learning remains sensitive to estimation noise during action selection.
Fundamentally, deep Q-learning is a stochastic regression problem: 
the value function is optimized to minimize a squared-error objective with bootstrapped targets.
Although such regression may be unbiased in expectation, control decisions rely on the maximization of noisy value estimates.

Suppose the learned action-value function satisfies
\begin{equation}
Q_\theta(s,a) = Q^*(s,a) + \varepsilon_a,
\end{equation}
where $\varepsilon_a$ is a zero-mean estimation error with 
$\mathbb{E}[\varepsilon_a] = 0$.
The greedy policy selects
\begin{equation}
a_\theta^*(s) = \arg\max_a Q_\theta(s,a).
\end{equation}
Let $\Delta(s)$ denote the action gap between the optimal action $a^*$ and the second-best alternative:
\begin{equation}
\Delta(s) = Q^*(s,a^*) - \max_{a\neq a^*} Q^*(s,a).
\end{equation}
The probability of selecting a suboptimal action is therefore
\begin{equation}
\Pr(a_\theta^*(s) \neq a^*) 
= \Pr(\varepsilon_{a'} - \varepsilon_{a^*} > \Delta(s)),
\end{equation}
for some $a' \neq a^*$.

Thus, even unbiased estimation noise can induce unstable action selection when the action gap is small.
While this does not introduce systematic value bias, it directly affects the quality of decisions made during interaction with the environment.
Because the behavior policy determines the distribution of collected data, 
noisy action selection may lead the agent to generate lower-return trajectories.
Learning slows not due to inefficient reuse of samples, but because the interaction policy itself produces less informative experience.
In this sense, estimator-level instability degrades control reliability and indirectly limits performance improvement.

These observations motivate the need for variance-reduction techniques in value regression, so that action selection remains stable under stochastic estimation noise.

\subsection{Parameter-Dynamics Instability: Layer-wise Parameter Imbalance and Plasticity Loss\label{sec: param_level}}
Beyond operator-level bias and estimator-level decision sensitivity, 
instability may also arise from the evolution of parameters under non-stationary sampling.

In reinforcement learning, the data distribution is determined by the behavior policy, 
which continuously changes throughout training.
As a result, the representation learned by the network must remain adaptable to shifting state-action distributions.
However, repeated optimization on a limited or slowly evolving data distribution can progressively bias parameter updates toward frequently observed patterns.

This effect exists in general, but becomes increasingly pronounced as the replay ratio grows.
When past experiences are reused many times before sufficient new interactions are collected, the influence of the current data distribution is amplified.
Over time, this can drive certain weights to dominate their layers, while the remaining weights receive comparatively weaker updates.
The resulting parameter distribution becomes increasingly skewed.

To quantify this imbalance, we introduce a diagnostic statistic termed the spike ratio.
For a given layer $\ell$ with parameters $\theta_\ell$, we define
\begin{equation}
\label{equ: spike_ratio}
\mathrm{SpikeRatio}_\ell
=
\frac{\|\theta_\ell\|_\infty}
{\mathrm{Quantile}_{0.99}(|\theta_\ell|)}.
\end{equation}
It measures the dominance of extreme parameter values relative to the bulk of the distribution.
An increasing spike ratio indicates that a small subset of weights grows disproportionately large, revealing an imbalanced representation.

\begin{figure}[!t]
\centering
\includegraphics[width=\linewidth]{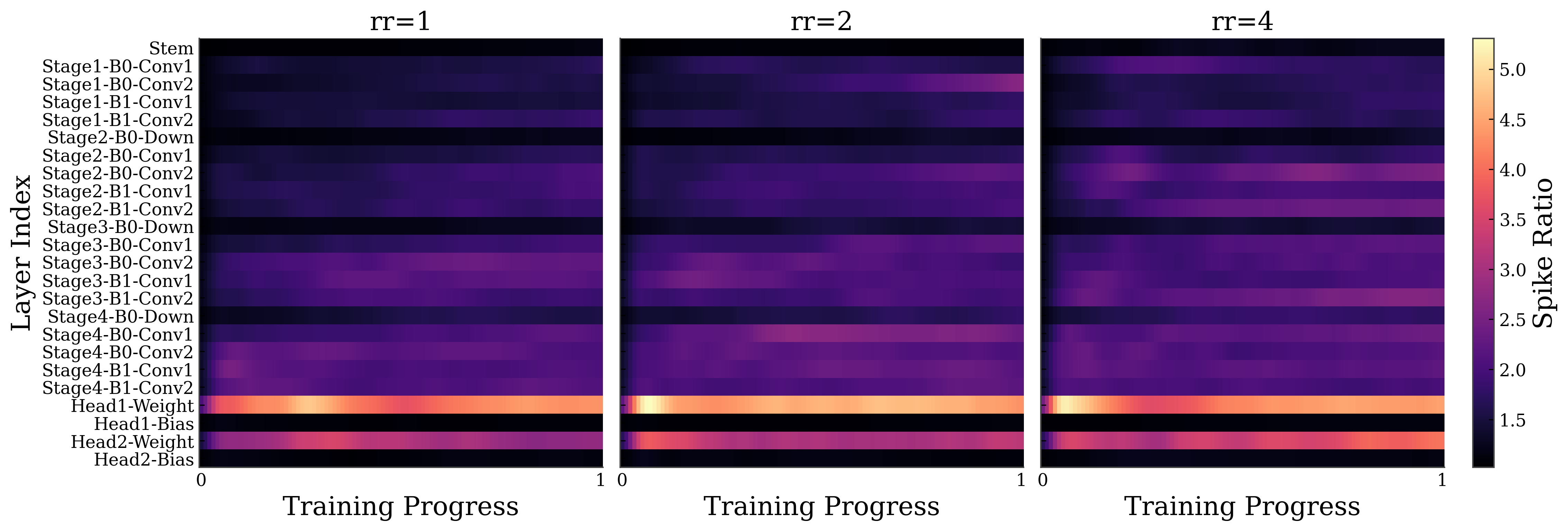}
\caption{Layer-wise spike ratio evolution during training under different replay ratios. Each row corresponds to a network layer, and color intensity indicates spike magnitude. Higher replay ratios lead to more pronounced spike growth, particularly in post-encoder MLP layers. Further details are provided in the supplementary materials.}
\label{fig: ab_spikeratio}
\end{figure}
Fig.~\ref{fig: ab_spikeratio} illustrates the evolution of $\mathrm{SpikeRatio}_\ell$ under different replay ratios ($\mathrm{rr}=1,2,4$).
While spike growth is observable even at lower replay ratios, it becomes more pronounced as the replay ratio increases, particularly in post-encoder MLP layers. This pattern indicates that replay amplification systematically accelerates parameter imbalance.

Parameter-dynamics instability does not necessarily lead to numerical divergence of value estimates.
Instead, it reflects a gradual loss of representational plasticity.
As the data distribution shifts with policy improvement,
a skewed parameter configuration may adapt more slowly,
thereby constraining long-term performance improvement.
These observations underscore the importance of preserving balanced parameter evolution throughout training.

\section{Stabilization Principles for Value Learning\label{sec:method}}

The analysis in Section~\ref{sec:analyse} reveals that instability in deep Q-learning arises from three interacting mechanisms: operator-level bias in recursive bootstrapping, estimator-level sensitivity of greedy action selection to regression noise, and parameter-dynamics imbalance under non-stationary sampling.
These mechanisms interact through the recursive structure of value learning. Instability may emerge when bootstrap targets amplify value estimates, when noisy regression alters control decisions, or when network parameters gradually lose plasticity during prolonged training. Stabilizing value learning therefore requires coordinated regulation of these components rather than isolated heuristics.

Based on the preceding analysis, we derive three stabilization principles for deep Q-learning:

\begin{itemize}
\item \textbf{Controlled bootstrapping.} 
Recursive Bellman updates must be regulated to prevent feedback amplification during target construction.

\item \textbf{Variance-aware value estimation.} 
Reducing estimation variance improves the reliability of greedy action selection and stabilizes interaction dynamics.

\item \textbf{Parameter-dynamics regulation.} 
Maintaining balanced parameter evolution preserves network plasticity under prolonged replay.
\end{itemize}

In the following, we describe practical mechanisms that instantiate these principles. Together they form a controlled optimization procedure for stabilizing value learning.

\subsection{Controlled Bootstrapping}
Section~\ref{sec: operator_level} shows that operator-level instability arises not only from classical maximization bias, but also from a previously overlooked reward-triggered self-reinforcing pathway in recursive bootstrapping.

\subsubsection{Cross-Model Decoupling for Action Selection and Evaluation}
The overestimation phenomenon in Eq.~\eqref{equ: decouple} arises from using the same noisy estimator for both action selection and value evaluation. Double DQN alleviates this issue by decoupling these roles using online and target networks.

We generalize this idea to an ensemble setting. Let
\begin{equation}
Q = \{\hat{Q}_1, \ldots, \hat{Q}_m\}
\end{equation}
denote an ensemble of value functions. For each model $\hat{Q}_i$, we select the bootstrap action using another model $\hat{Q}_j$ ($j \neq i$), while the value is evaluated using $\hat{Q}_i$:

\begin{equation}
y_i = r + \gamma \hat{Q}_i \left( s', \arg\max_{a'} \hat{Q}_j(s',a') \right).
\end{equation}

This structurally eliminates selection-evaluation coupling.
When the estimation noises of $\hat{Q}_i$ and $\hat{Q}_j$ are weakly correlated as encouraged by independent initialization and separate bootstrap targets, the selection noise and evaluation noise are effectively decoupled, which substantially reduces the upward bias induced by the max operator.

\subsubsection{Action-Decoupling for Reward-Bearing Transitions}

The instability analysis in Section~\ref{sec: operator_level} further reveals a feedback pathway termed the self-reinforcing trap. When a reward-bearing transition $(s,a,r,s')$ is repeatedly replayed, representation generalization may increase both $Q(s,a)$ and $Q(s',a)$ simultaneously, making action $a$ likely to be selected again during the bootstrap step.

To break this feedback loop, we impose a constraint on reward-bearing transitions:

\begin{equation}
a'_i \neq a \quad \text{if } r>0.
\end{equation}

This constraint prevents the bootstrap target from repeatedly selecting the same action that produced the reward, thereby disrupting the amplification pathway identified in the instability analysis.

\subsubsection{Bounded Bellman Updates}

Even with decoupled bootstrapping, recursive updates may gradually increase the numerical scale of value estimates. To prevent uncontrolled growth, we impose an upper bound on target values derived from the discounted return along greedy trajectories.

Given a greedy trajectory $\mathcal{T}$, we define a trajectory-level bound
\begin{equation}
B(\mathcal{T}) = \max_{t} G_t,
\end{equation}
where $G_t$ denotes the discounted return. This bound provides a natural constraint on the numerical scale of value estimates during training.
Further implementation details are provided in the supplementary material.

\subsection{Variance-Aware Value Estimation}

Section~\ref{sec: estimator_level} shows that even unbiased regression noise may alter greedy action selection when action gaps are small. This effect may destabilize interaction dynamics.

To reduce estimation variance, we adopt ensemble quantile regression. Following QR-DQN, we model the return distribution using $K$ quantile estimates

\begin{equation}
Z_\theta(s,a) = \frac{1}{K} \sum_{i=1}^{K} \delta_{z_\theta^{(i)}(s,a)}.
\end{equation}
The scalar value estimate is
\begin{equation}
\label{equ: quantile_q}
Q_\theta(s,a) = \frac{1}{K} \sum_{i=1}^{K} z_\theta^{(i)}(s,a).
\end{equation}
Given bootstrap action $a'^*$, the target quantiles are
\begin{equation}
y^{(j)} = r + \gamma z_{\bar{\theta}}^{(j)}(s',a'^*),
\end{equation}
and parameters are optimized via the quantile regression loss
\begin{equation}
\label{equ: loss}
\mathcal{L}_{\text{QR}}(\theta)
=
\frac{1}{K^2}
\sum_{i=1}^{K}
\sum_{j=1}^{K}
\rho_{\tau_i}^{\kappa}\!\left(y^{(j)} - z_\theta^{(i)}(s,a)\right).
\end{equation}

To further reduce variance, we maintain an ensemble of quantile networks
\begin{equation}
\{Z_{\theta_1},\ldots,Z_{\theta_m}\}.
\end{equation}
The ensemble-averaged value used for action selection is
\begin{equation}
\label{equ: ensemble_q}
\bar{Q}(s,a)=\frac{1}{m}\sum_{\ell=1}^{m}\frac{1}{K}\sum_{i=1}^{K}z^{(i)}_{\theta_\ell}(s,a).
\end{equation}
Ensemble aggregation reduces both intra-model and inter-model estimation variance, improving the reliability of greedy action selection.

\subsection{Parameter-Dynamics Regulation}

Section~\ref{sec: param_level} reveals that high replay ratios may gradually distort parameter distributions, leading to reduced network plasticity.
To monitor this phenomenon, we introduce the spike ratio, defined as Eq.~\eqref{equ: spike_ratio}.
This statistic measures the dominance of extreme parameter values relative to the bulk of the distribution.
When the spike ratio of a layer exceeds a predefined threshold, corrective intervention is triggered by resetting the affected parameters. This mechanism restores distributional balance and helps preserve long-term adaptability of the value network.

\section{Algorithm Instantiation\label{sec: algo}}
The mechanisms described above provide practical implementations of the stabilization principles derived from the instability analysis. Together they form a controlled optimization procedure for value learning.
\begin{algorithm}[!t]
\caption{Bootstrapping Control and Ensemble Quantile Regression}
\label{algo: eqrdqn}
\SetAlgoLined
\DontPrintSemicolon
\KwIn{
Replay buffer $\mathcal{D}$; ensemble size $m$; number of quantiles $K$;
Online networks $\{Q_i(\cdot;\theta_i)\}_{i=1}^m$ and target networks $\{Q_i(\cdot;\theta_i^-)\}_{i=1}^m$;
Quantile fractions $\{\tau_j = \tfrac{j-0.5}{K}\}_{j=1}^K$; exploration rate $\epsilon$
}
\KwOut{Updated ensemble parameters $\{\theta_i\}_{i=1}^m$}

\BlankLine
\For{$t=1$ \KwTo $T$}{

    \tcp{Environment interaction}
    Observe state $s_t$\;
    \eIf{$\mathrm{rand()} < \epsilon$}{
        Select $a_t \sim \mathrm{Uniform}(\mathcal{A})$\;
    }{
        Select greedy action according to ensemble-averaged value(Eq.~\eqref{equ: ensemble_q}):
        $a_t = \arg\max_{a\in\mathcal{A}} \bar{Q}(s_t,a)$
    }
    Execute $a_t$, observe $(r_t, s_{t+1})$, and store $(s_t,a_t,r_t,s_{t+1})$ in $\mathcal{D}$\;

    \BlankLine
    \tcp{Prioritized experience replay}
    Sample a mini-batch $\{(s^{(b)},a^{(b)},r^{(b)},s'^{(b)})\}_{b=1}^{B}$ from $\mathcal{D}$\;

    \tcp{Bootstrap action selection}
    \For{$i=1$ \KwTo $m$}{
        Aggregate distributional value estimation according to Eq.~\eqref{equ: quantile_q}\;
        Decoupling reward-bearing transitions:\;
        \If{$r^{(b)}>0$}{
            $Q_{\theta_i^-}(s'^{(b)},a^{(b)})=-\infty$
        }
        Select bootstrap action:
        $a_i^{*} = \arg\max_{a'\in\mathcal{A}} Q_{\theta_i^-}(s',a')$
    }

    Sample a random permutation $\pi$ over $\{1,\dots,m\}$\;

    \For{$i=1$ \KwTo $m$}{
        \tcp{Cross-model bootstrapping}
        \For{$j'=1$ \KwTo $K$}{
            Compute target quantiles: $y_i^{(b,j')} = r^{(b)} + \gamma\,z_{i,j'}\!\left(s'^{(b)}, a^{*}_{\pi(i)};\theta_i^- \right)$
        }
        Compute quantile regression loss and update parameters according to Eq.~\eqref{equ: loss}\;
    }

    Periodically update target networks $\theta_i^- \leftarrow \theta_i$\;
    Periodically monitor layer-wise spike ratios according to Eq.~\eqref{equ: spike_ratio}\;
    \If{$\mathrm{Layer}\ \ell\mathrm{'s\ spike\ ratio\ is\ greater\ than\ threshold}$}{
        Reset layer $\ell$'s parameters
    }
}
\end{algorithm}

Algorithm~\ref{algo: eqrdqn} summarizes the overall training process.
At each interaction step (Lines~1--8), the agent observes the current state $s_t$ (Line~2) and selects an action using $\epsilon$-greedy exploration (Lines~3--7). With probability $\epsilon$, a random action is sampled uniformly (Line~4). Otherwise, the agent acts greedily w.r.t.\ the ensemble-averaged value $\bar{Q}(s,a)$ defined in Eq.~\eqref{equ: ensemble_q} (Line~6).
The resulting transition $(s_t,a_t,r_t,s_{t+1})$ is then stored in the replay buffer $\mathcal{D}$ (Line~8).
For learning, we sample a prioritized minibatch $\{(s^{(b)},a^{(b)},r^{(b)},s'^{(b)})\}_{b=1}^{B}$ from $\mathcal{D}$ (Line~9).
Next, we determine bootstrap actions for each ensemble member using its own target network (Lines~10--17).
Specifically, for each member $i$, we first form the scalar action values from its quantile outputs (Line~11, Eq.~\eqref{equ: quantile_q}) and then apply the reward-bearing decoupling rule (Lines~12--15):
for any sampled transition $b$ with $r^{(b)}>0$, we mask the self-reinforcing candidate by setting $Q_{\theta_i^-}(s'^{(b)},a^{(b)})=-\infty$, which enforces $a_i^*(s'^{(b)})\neq a^{(b)}$ during maximization.
We then compute the greedy bootstrap action
$a_i^*(s'^{(b)})=\arg\max_{a'\in\mathcal{A}} Q_{\theta_i^-}(s'^{(b)},a')$
(Line~16).
To decouple action selection from value evaluation across models, we sample a random permutation $\pi$ over ensemble indices (Line~18).
During the subsequent update (Lines~19--24), member $i$ evaluates the bootstrap action selected by another member $\pi(i)$.
Concretely, for each quantile index $j'\in\{1,\dots,K\}$, we construct the target quantile
$y_i^{(b,j')}=r^{(b)}+\gamma\, z_{i,j'}(s'^{(b)},a_{\pi(i)}^*(s'^{(b)});\theta_i^-)$
(Line~21), i.e., the action comes from the permuted selector while the quantile evaluation uses member $i$'s target network.
Member $i$ is then updated by minimizing the quantile regression objective in Eq.~\eqref{equ: loss} (Line~23).
Finally, target networks are periodically synchronized (Line~25), and we monitor layer-wise spike ratios (Line~26, Eq.~\eqref{equ: spike_ratio}) to detect over-adaptation.
If a layer's spike ratio exceeds a predefined threshold, we reset that layer's parameters to restore training stability (Lines~27--29).
\section{Experiments\label{sec: exp}}
% The experiments are designed to evaluate the proposed analysis and the stabilization principles derived from it.
Experiments evaluate the proposed analysis and the derived stabilization principles. 
First, we evaluate whether the algorithm instantiated from the proposed principles achieves competitive performance on standard reinforcement learning benchmarks. 
Then, we analyze how the proposed stabilization principles influence learning behavior through controlled ablations.

We begin with Atari-100K~\cite{ceron2021revisiting}(Section~\ref{sec: atari}), a widely adopted benchmark for data-efficient reinforcement learning. 
Due to its limited interaction budget and high replay utilization, Atari-100K provides a challenging setting where bootstrapping instability and value overestimation are particularly pronounced. 
This benchmark allows us to directly assess whether the proposed stabilization mechanisms improve learning reliability under constrained data conditions.

To further evaluate robustness in discrete control, we consider Procgen~\cite{cobbe2020leveraging}(Section~\ref{sec: procgen}), which features procedurally generated environments with diverse visual appearances and dynamics. 
Unlike fixed-layout benchmarks, Procgen tests representation robustness and generalization across unseen levels, enabling us to examine whether improved value stability translates into stronger cross-distribution performance.

In addition to benchmark comparisons, we perform a series of diagnostic analyses to investigate the mechanisms identified in Section~\ref{sec:analyse} and the stabilization principles introduced in Section~\ref{sec:method}.

\subsection{Atari-100K: Learning Stability under Data-Efficient Regimes\label{sec: atari}}
We evaluate our method on the Atari-100K benchmark, which consists of 26 Atari 2600 games under a strict interaction budget of 100k environment steps.
This corresponds to 400k frames with a frame skip of 4.
% Under such a limited data regime, agents must repeatedly reuse collected transitions during training, typically resulting in high replay ratios.
Under this limited data regime, agents must repeatedly reuse collected transitions, resulting in high replay ratios.

Our implementation follows the standard Atari preprocessing pipeline~\cite{mnih2015human}.
Observations are resized and stacked following common practice.
We adopt a convolutional ResNet-style backbone without normalization layers, consistent with our method design.
Results are averaged over multiple random seeds.
Detailed hyperparameter configurations are provided in supplementary materials.

We compare against a diverse set of representative methods evaluated under the Atari-100K protocol.
Random and human scores are reported as reference points.
DER~\cite{van2019use} incorporates auxiliary supervised objectives into value learning to improve sample efficiency.
DrQ~\cite{yarats2021image} enhances data efficiency via image augmentation applied directly to the value function.
IRIS~\cite{micheli2022transformers}, STORM~\cite{zhang2023storm}, DreamerV3~\cite{hafner2023mastering}, DIAMOND~\cite{alonso2024diffusion}, DART~\cite{agarwal2024learning}, and Drama~\cite{wang2024drama} are model-based or hybrid approaches that leverage latent dynamics modeling for improved planning and sample efficiency.
REM~\cite{agarwal2020optimistic} reduces overestimation through ensemble-based value aggregation.
SGF~\cite{robine2025simple} focuses on stabilizing gradient propagation during representation learning.
BBF~\cite{schwarzer2023bigger} combines aggressive replay, regularization, and parameter resetting to achieve strong performance under limited data.
These baselines collectively cover value-based, model-based, and hybrid paradigms for data-efficient reinforcement learning.

Performance is measured using the Human-Normalized Score (HNS).
% computed per game as
% \[
% \text{HNS} = \frac{\text{Score}_{\text{agent}} - \text{Score}_{\text{random}}}
% {\text{Score}_{\text{human}} - \text{Score}_{\text{random}}}.
% \]
We report aggregate statistics across 26 games, including the mean HNS, median HNS, and Interquartile Mean (IQM), which provides a robust estimate that is less sensitive to extreme outliers.
We additionally report the number of games surpassing human-level performance and the number of per-game best results.

\begin{table}[!t]
\centering
\small
\caption{
Aggregate performance on Atari-100K across 26 games.
All metrics are computed using human-normalized scores (HNS).
Mean, Median, and Interquartile Mean (IQM) are reported following standard evaluation practice.
\#Human denotes the number of games surpassing human-level performance,
and \#Best indicates the number of per-game best results.
Higher is better for all metrics.
}
\label{tab: atari}
\begin{tabularx}{\columnwidth}{
l
>{\centering\arraybackslash}X
>{\centering\arraybackslash}X
>{\centering\arraybackslash}X
>{\centering\arraybackslash}X
>{\centering\arraybackslash}X
}
\toprule
Algorithm
& \#Human
& \#Best
& Mean
& Median
& IQM\\
\midrule
DER        & 2  & 0  & 0.350 & 0.189 & 0.183 \\
DrQ        & 3  & 0  & 0.465 & 0.313 & 0.280 \\
IRIS       & 9  & 0  & 1.046 & 0.289 & 0.501 \\
REM        & 12 & 1  & 1.222 & 0.280 & 0.673 \\
STORM      & 10 & 5  & 1.266 & 0.580 & 0.636 \\
DreamerV3  & 9  & 1  & 1.120 & 0.466 & 0.490 \\
DIAMOND    & 11 & 0  & 1.459 & 0.373 & 0.641 \\
DART       & 9  & 0  & 1.022 & 0.790 & 0.575 \\
SGF        & 6  & 0  & 0.884 & 0.152 & 0.287 \\
Drama      & 8  & 1  & 1.049 & 0.270 & 0.367 \\
BBF        & 12 & 8  & \textbf{2.247} & 0.917 & 1.045 \\
Ours       & \textbf{14} & \textbf{10} & 1.799 & \textbf{1.045} & \textbf{1.070} \\
\bottomrule
\end{tabularx}
\end{table}
Aggregate results are presented in Table~\ref{tab: atari}.
Our method achieves the highest IQM and Median HNS among all compared approaches.
We also obtain the largest number of games exceeding human-level performance and the highest number of per-game best results.
These results indicate strong overall performance under the stringent 100k interaction constraint. For completeness and transparency, full per-environment scores are reported in supplementary materials.

\subsection{Procgen: Representation Robustness and Generalization\label{sec: procgen}}
We further evaluate our method on the Procgen benchmark~\cite{cobbe2020leveraging}, which emphasizes representation robustness and cross-level generalization.
Unlike Atari-100K, where training and evaluation share identical game layouts,
Procgen generates diverse levels procedurally and allows strict separation 
between training and test environments.

We adopt the generalization protocol, where the agent is trained on
a fixed set of 200 procedurally generated levels and evaluated on an unseen
set of test levels.
This setting explicitly measures cross-level generalization ability rather
than memorization of specific layouts.
All agents interact with environments for a total of 25M environment steps.
During evaluation, exploration is disabled and performance is averaged over
multiple test levels.
Detailed hyperparameter configurations are provided in supplementary materials.

We compare against the official PPO implementation reported in the Procgen paper~\cite{cobbe2020leveraging}.
PPO is widely adopted as a strong on-policy baseline for Procgen and has
been carefully tuned for this benchmark.
Using PPO as the primary baseline allows us to assess whether improved
value stability translates into stronger generalization under procedural diversity.

For each game, we report the average test return over unseen levels.
To summarize performance across environments, we additionally report:
the mean score, the median score, and the interquartile mean (IQM),
following recent evaluation practice for robust aggregate comparison.
Higher values indicate better generalization performance.

\begin{table}[t]
\centering
\caption{Procgen (easy difficulty, 200 training levels). 
Scores are normalized using min-max normalization as defined in the original Procgen paper.}
\label{tab: procgen}
\begin{tabularx}{\columnwidth}{
l
>{\centering\arraybackslash}X
>{\centering\arraybackslash}X
>{\centering\arraybackslash}X
}
\toprule
 & Test Mean & Test Median & Test IQM \\
\midrule
PPO (Impala CNN) & 0.33 & 0.38 & 0.27 \\
Ours & 0.40 & 0.40 & 0.39 \\
\bottomrule
\end{tabularx}
\end{table}
Table~\ref{tab: procgen} presents aggregate statistics.
Our method consistently outperforms the baseline on the majority of environments, achieving higher average scores.
The improvement is particularly pronounced in environments with high visual and structural variability, suggesting that stabilizing recursive value learning benefits representation robustness under procedural diversity. Detailed results and additional visualization are provided in the supplementary materials.

% A closer examination of per-environment performance reveals a structured pattern rather than uniform improvement.
Per-environment performance reveals a structured pattern rather than uniform improvement.

Substantial gains are observed in environments such as \emph{BigFish}, \emph{Dodgeball}, and \emph{StarPilot}, which provide relatively dense reward signals and require stable action-value propagation.
In these settings, recursive bootstrapping occurs frequently, and over-amplified value estimates can easily destabilize training. By controlling bootstrap bias and reducing estimation variance, our method produces smoother value propagation and more reliable action selection, leading to significant performance improvements.

In contrast, environments such as \emph{Chaser}, \emph{Heist}, and \emph{Maze} remain challenging.
These tasks feature sparse rewards and long-horizon credit assignment, where performance is heavily influenced by exploration efficiency and global planning capability.
Since our method primarily targets value instability rather than exploration enhancement, its advantages are less pronounced in such environments.

\subsection{Mechanism Analysis and Sensitivity Study\label{sec: analysis_exp}}
Beyond benchmark results, we conduct additional experiments to analyze the behavior of the proposed method.
Specifically, we examine the effects of ensemble size, the Action-Decoupling Constraint, parameter-level spike dynamics, and replay ratio.
These studies aim to provide further insight into the factors influencing training stability and performance.

\subsubsection{Effect of Ensemble Size}
We investigate the influence of the number of ensemble networks $N_e$ on training performance.
Experiments are conducted on four representative Atari environments: \textit{Alien}, \textit{Amidar}, \textit{BankHeist}, and \textit{Breakout}.
We evaluate $N_e \in \{2,4,8,16\}$ while keeping all other training settings fixed.
Each configuration is repeated with multiple random seeds.

\begin{figure}[!t]
\centering
\includegraphics[width=\linewidth]{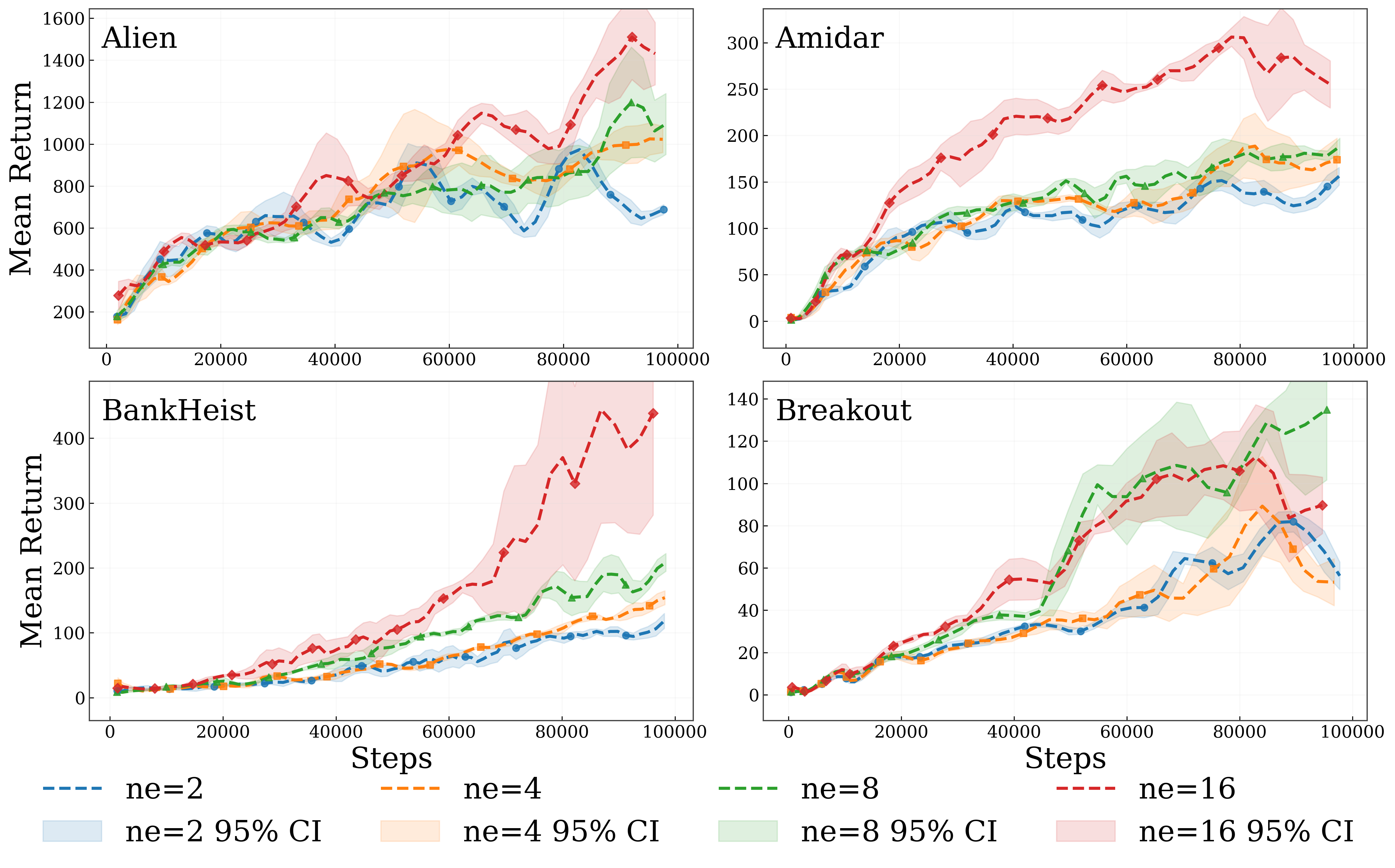}
\caption{Training performance under different ensemble sizes $N_e$ on four Atari-100K environments. Curves show the recent 10-episode mean return; shaded regions denote 95\% CIs.}
\label{fig: ab_ensemble}
\end{figure}
Fig.~\ref{fig: ab_ensemble} shows the evolution of the recent 10-episode mean return during training.
Increasing the ensemble size generally improves performance and training stability across most environments.
In particular, larger ensembles lead to faster performance growth and higher final returns in \textit{Alien}, \textit{Amidar}, and \textit{BankHeist}.
Breakout exhibits a less monotonic trend, indicating that the effect of ensemble size may vary depending on environment dynamics.

\begin{figure}[!t]
\centering
\includegraphics[width=\linewidth]{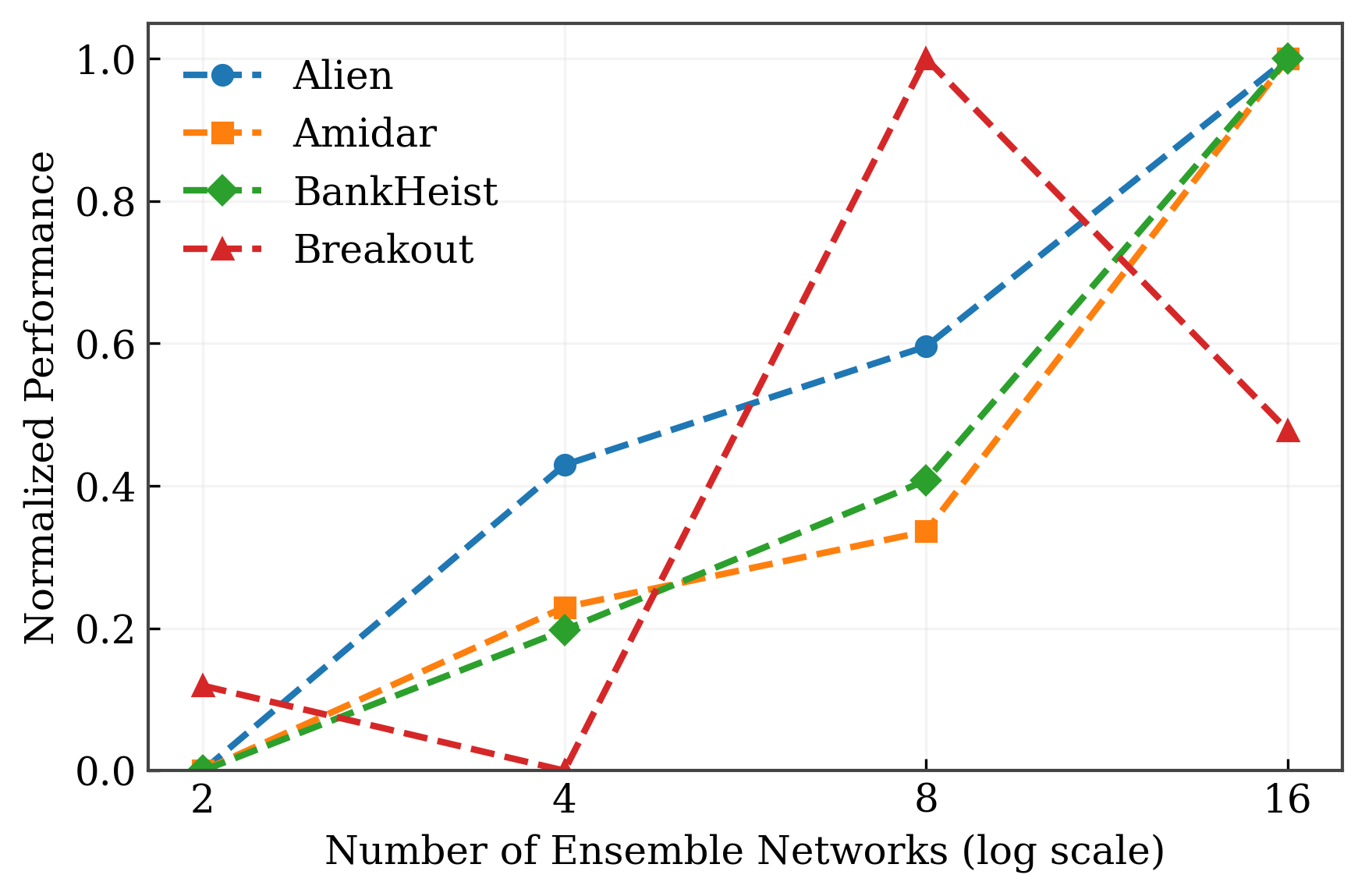}
\caption{Normalized final performance as a function of the ensemble size. Results are normalized within each environment.}
\label{fig: ab_ensemble_summary}
\end{figure}
To further summarize the overall trend, Fig.~\ref{fig: ab_ensemble_summary} reports normalized final performance as a function of $N_e$.
Performance improves consistently as the ensemble size increases from 2 to 16 in most cases, suggesting that larger ensembles contribute positively to learning stability.

\subsubsection{Effect of the Action-Decoupling Constraint}
We evaluate the impact of the Action-Decoupling Constraint applied to reward-bearing transitions by comparing the full method with a variant in which the constraint is removed during target computation.
All other training settings are kept identical.
Experiments are conducted on \textit{Alien}, \textit{Amidar}, \textit{BankHeist}, and \textit{Breakout}.

\begin{figure}[!t]
\centering
\includegraphics[width=\linewidth]{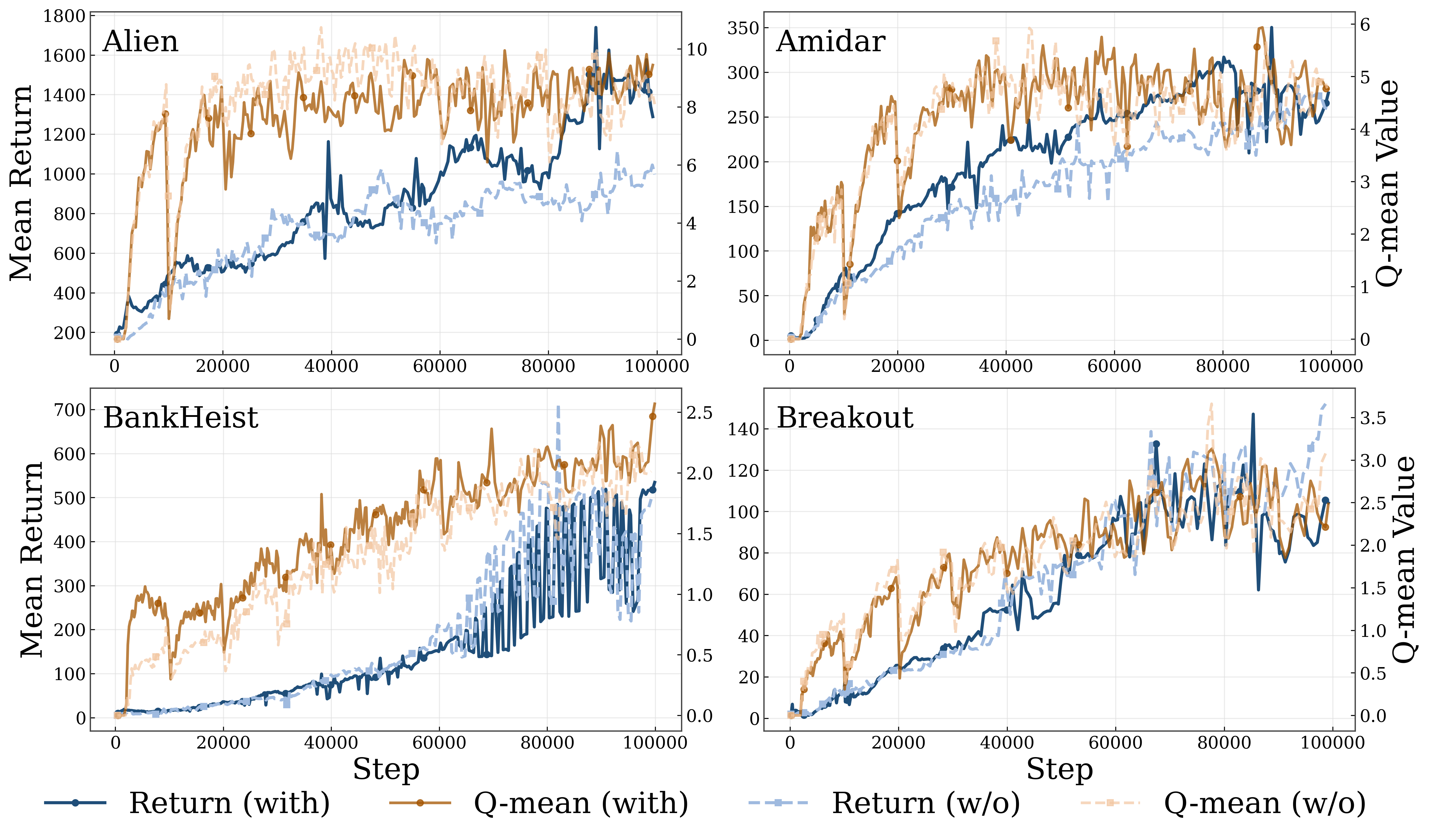}
\caption{Comparison of training dynamics with and without the Action-Decoupling Constraint. For each environment, the left axis shows the recent 10-episode mean return, and the right axis shows the average Q-value. Solid lines correspond to the constrained version, while dashed lines denote the unconstrained variant.}
\label{fig: ab_decouple}
\end{figure}
Fig.~\ref{fig: ab_decouple} presents the evolution of the recent 10-episode mean return (left axis) and the average Q-value (right axis).
On \textit{Alien}, the constrained version yields a clear and consistent performance improvement over the unconstrained variant.
In contrast, on \textit{Amidar}, \textit{BankHeist}, and \textit{Breakout}, the performance differences are less pronounced, and in some cases the unconstrained version performs comparably or slightly better.

A key distinction lies in reward density.
\textit{Alien} provides relatively dense and frequent rewards, where excessive self-reinforcing updates may amplify overestimated values and destabilize value learning.
The Action-Decoupling Constraint mitigates this amplification, leading to more stable performance.
In environments with sparser rewards, moderate self-reinforcing effects can help strengthen weak learning signals, which may explain the smaller performance gap in those tasks.
In such cases, weaker signal amplification can also be compensated by longer training or increased data reuse.
Further analysis under higher replay ratios is provided later in this section.

\subsubsection{Parameter-Level Spike Dynamics and Reset Mechanism}
We first analyze parameter-level spike dynamics during training.
For each network layer, we monitor the spike ratio (Eq.~\eqref{equ: spike_ratio}) over training progress, defined as the ratio between the maximum absolute parameter value and a high-percentile statistic within the same layer.
This metric provides a proxy for detecting abnormal parameter amplification.

Fig.~\ref{fig: ab_spikeratio} visualizes the spike ratio evolution under different replay ratios ($\mathrm{rr}=1,2,4$).
Two consistent patterns emerge.
First, spike ratios are significantly higher in the MLP layers following the encoder, particularly in the projection head.
In contrast, convolutional encoder layers remain comparatively stable.
Second, increasing the replay ratio leads to systematically higher spike ratios across layers.
This suggests that aggressive data reuse amplifies parameter imbalance and increases the risk of unstable value growth.

We further investigate the role of the parameter reset mechanism.
Specifically, we compare training performance with and without parameter resetting, and record the frequency of reset events during training.
\begin{figure}[!t]
\centering
\includegraphics[width=\linewidth]{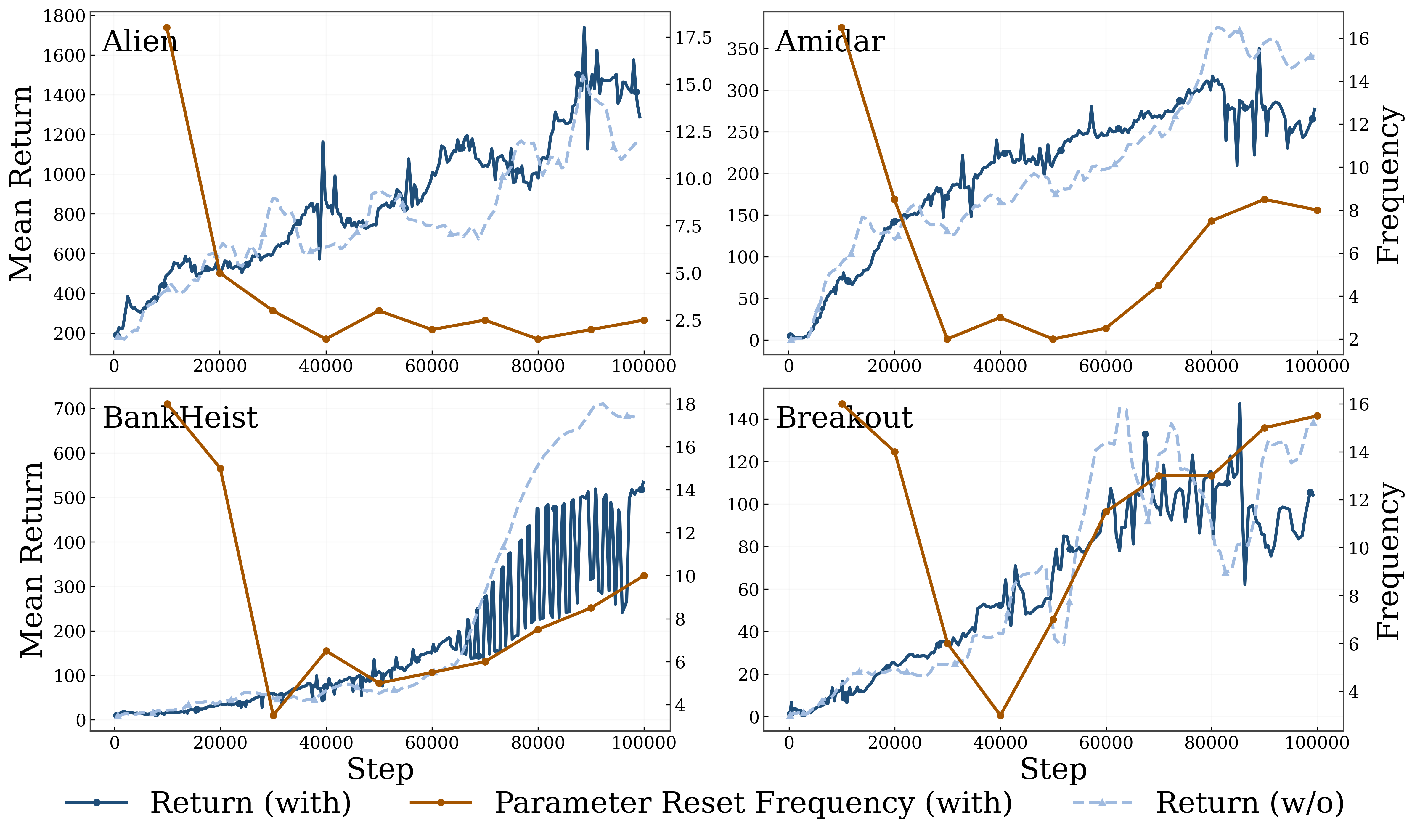}
\caption{Effect of parameter resetting on training performance.
Solid lines denote the version with parameter reset, while dashed lines indicate the variant without resetting. The right axis shows the frequency of reset events during training.}
\label{fig: ab_paramreset}
\end{figure}
As shown in Fig.~\ref{fig: ab_paramreset}, parameter resets occur more frequently in early training stages and under environments where value growth is more volatile.
When parameter resetting is enabled, its impact varies across environments.
Among the four evaluated tasks, a clear improvement is observed on \textit{Alien}, while the effect on the other environments is comparatively modest.
This suggests a trade-off: although resetting can restore network plasticity, it may also introduce irreversible performance loss by disrupting useful parameter structures that have already formed during training.
As learning progresses, the model may gradually specialize toward informative features, and aggressively preserving plasticity can therefore interfere with stable decision making.
Figure~2 further shows that, although plasticity tends to decrease during training, it can partially recover as new interaction data are incorporated, rather than collapsing permanently.
Empirically, higher replay ratios require stricter and more frequent parameter resetting to maintain stable training dynamics.

\subsubsection{Sensitivity to Replay Ratio}
We further evaluate the sensitivity of the proposed method to replay ratio (rr), which controls the degree of data reuse during training.
Experiments are conducted on \textit{Alien}, \textit{Amidar}, \textit{BankHeist}, and \textit{Breakout} with $\mathrm{rr} \in \{1,2,4,8\}$, while keeping all other hyperparameters fixed.

\begin{figure}[!t]
\centering
\includegraphics[width=\linewidth]{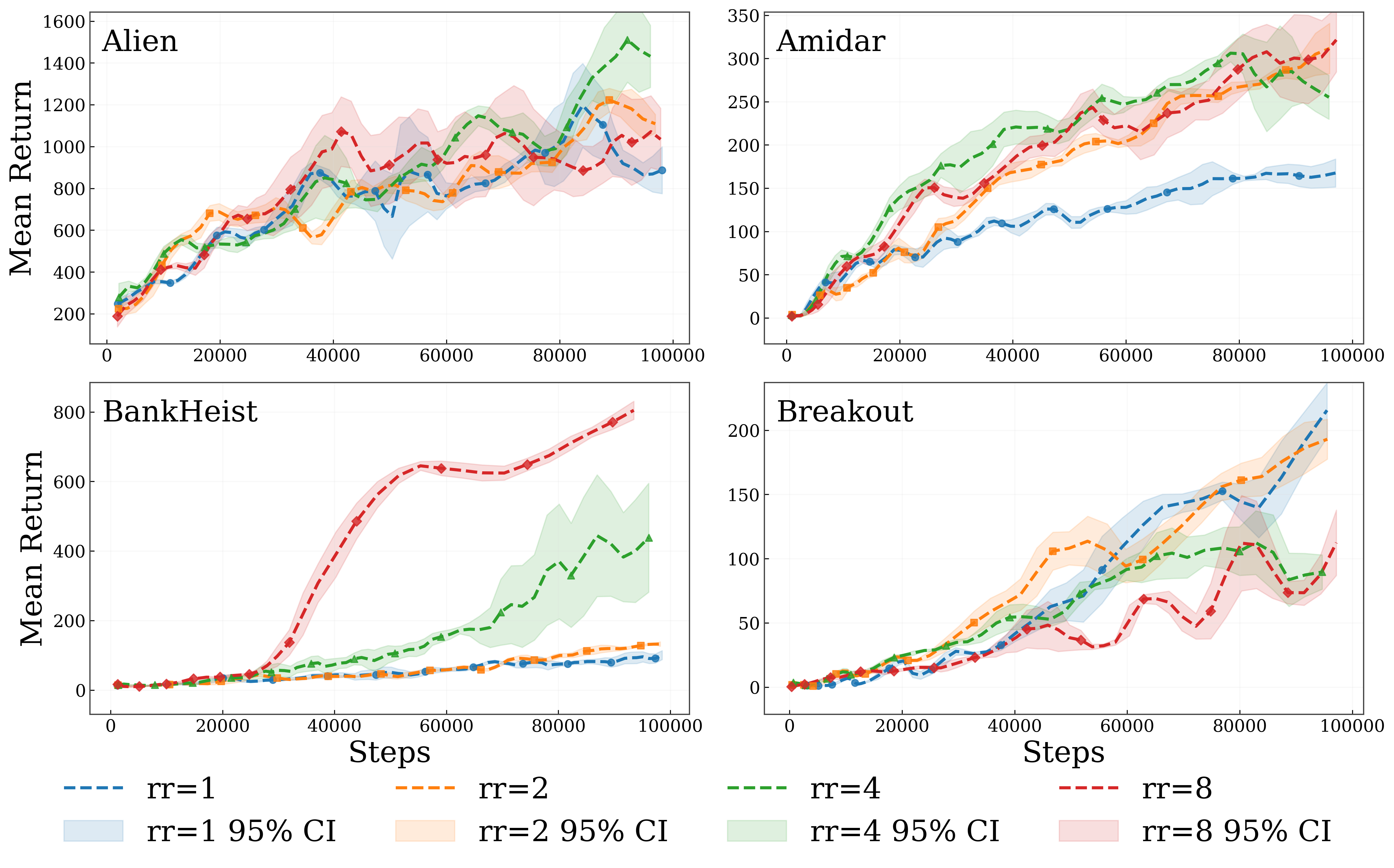}
\caption{Training performance under different replay ratios on four Atari-100K environments. Curves show the recent 10-episode mean return; shaded regions denote 95\% CIs.}
\label{fig: ab_rr}
\end{figure}
Fig.~\ref{fig: ab_rr} presents training curves under different replay ratios.
The impact of data reuse varies across environments.
In \textit{Alien} and \textit{Amidar}, moderate replay ratios (e.g., $\mathrm{rr}=4$) lead to improved performance compared to lower reuse settings.
However, excessively large replay ratios (e.g., $\mathrm{rr}=8$) may cause performance degradation in some environments.
In contrast, \textit{BankHeist} benefits consistently from higher replay ratios, suggesting that additional data reuse can help amplify weak learning signals in sparser reward settings.
Breakout exhibits a different trend, where performance peaks at lower replay ratios and decreases as data reuse becomes aggressive.

\begin{figure}[!t]
\centering
\includegraphics[width=\linewidth]{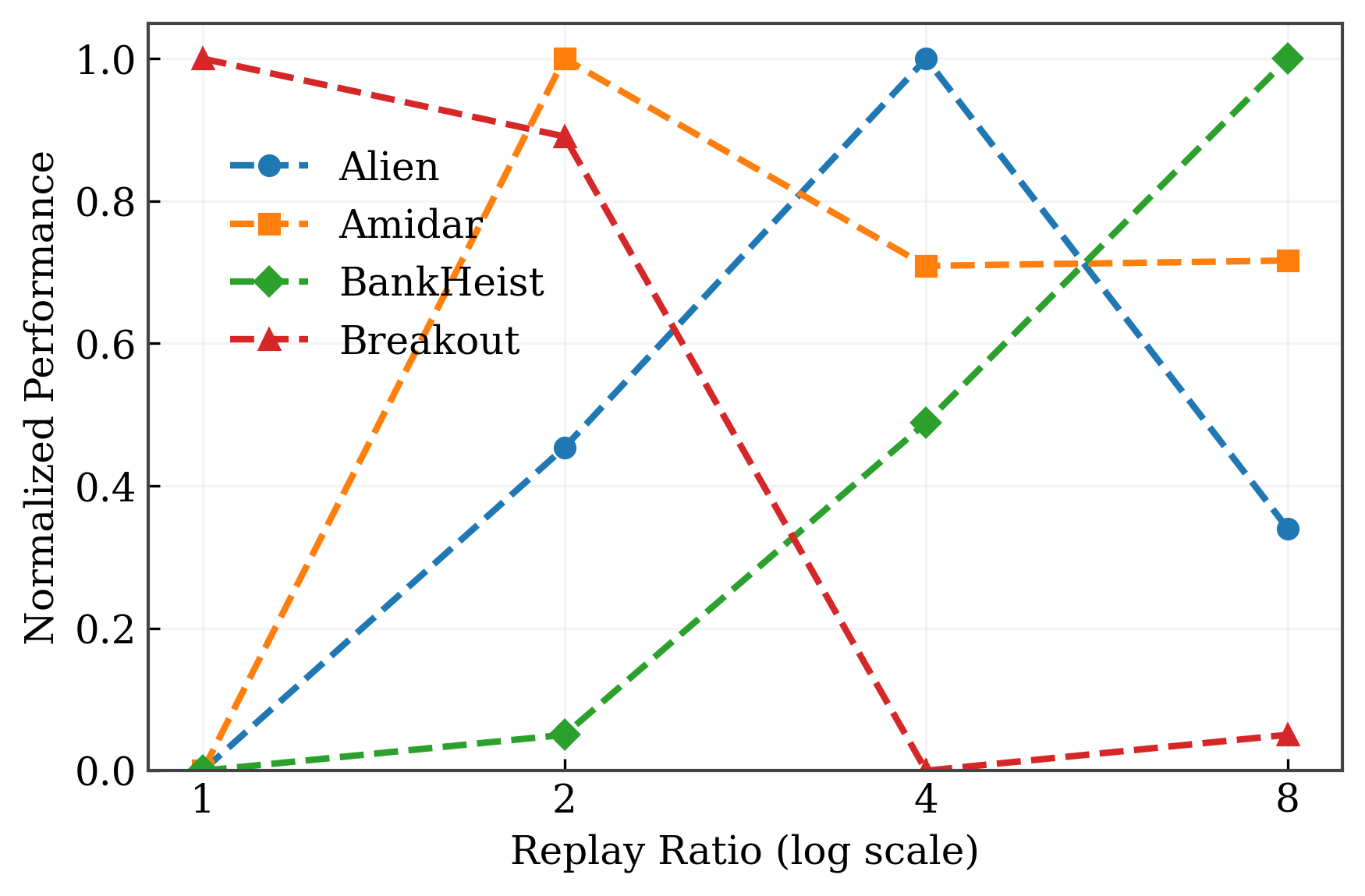}
\caption{Normalized final performance as a function of replay ratio. Results are normalized within each environment.}
\label{fig: ab_rr_summary}
\end{figure}
Fig.~\ref{fig: ab_rr_summary} summarizes the normalized final performance as a function of replay ratio.
The results reveal a non-monotonic relationship between replay ratio and final performance.
Across multiple environments, performance improves as $\mathrm{rr}$ increases from low to moderate values, but degrades when $\mathrm{rr}$ becomes excessively large.
This suggests that while additional data reuse enhances sample efficiency, there exists a task-dependent threshold beyond which further reuse no longer benefits learning.

A possible explanation relates to the interaction between aggressive data reuse and parameter spike dynamics.
As observed in the previous section, higher replay ratios systematically amplify spike ratios, especially in post-encoder layers.
Since spike monitoring and parameter resetting are performed periodically to avoid excessive computational overhead, very large replay ratios may lead to rapid over-amplification within a single monitoring interval.
In such cases, the network may temporarily overfit recent data before the next reset step is triggered, slowing down effective learning progress.
Consequently, excessively high replay ratios can reduce training efficiency.

These findings indicate that replay ratio must be carefully balanced with stability mechanisms, and that optimal data reuse is inherently environment-dependent.

\section{Conclusion and Future Work\label{sec: con}}
This work presents a unified perspective on instability in deep Q-learning by examining the dynamics of recursive value learning. 
We show that instability arises from the interaction of three mechanisms: operator-level bias in Bellman bootstrapping, estimator-level sensitivity of greedy decisions to regression noise, and parameter-dynamics imbalance under aggressive data reuse.

Our analysis reveals two characteristic behaviors that have received limited attention in prior studies. 
First, we identify a reward-triggered self-reinforcing trap in recursive bootstrapping, where value amplification interacts with representation generalization to repeatedly reinforce certain actions in bootstrap targets. 
Second, we uncover parameter spike dynamics associated with the gradual loss of network plasticity during training, and introduce the spike ratio as a practical diagnostic indicator for monitoring this effect.

Based on these insights, we derive a set of stabilization principles that regulate bootstrap target construction, reduce decision variance in value estimation, and maintain balanced parameter evolution during training. 
We instantiate these principles in a practical learning algorithm integrating controlled bootstrapping, ensemble quantile estimation, and spike-based parameter regulation.
Experiments on Atari-100K and Procgen demonstrate that the resulting method achieves competitive performance while improving training stability.

Beyond empirical results, our findings highlight the importance of analyzing reinforcement learning through training dynamics.
The interaction between bootstrapping, stochastic estimation, and parameter evolution suggests that stability in value learning is fundamentally a systems-level property rather than the result of isolated algorithmic components. 
Future work may explore adaptive stabilization strategies, automated coordination of hyperparameters such as replay ratio and ensemble size, and extensions of the proposed analysis to actor-critic methods and continuous-control settings.
\clearpage
\bibliographystyle{IEEEtran}
\bibliography{ref}

% The supplementary material is included in the same arXiv PDF in single-column format.
\clearpage
\onecolumn
\appendices
\section{Spike Ratio Dynamics}
\begin{figure}[h]
\centering
\includegraphics[width=\linewidth]{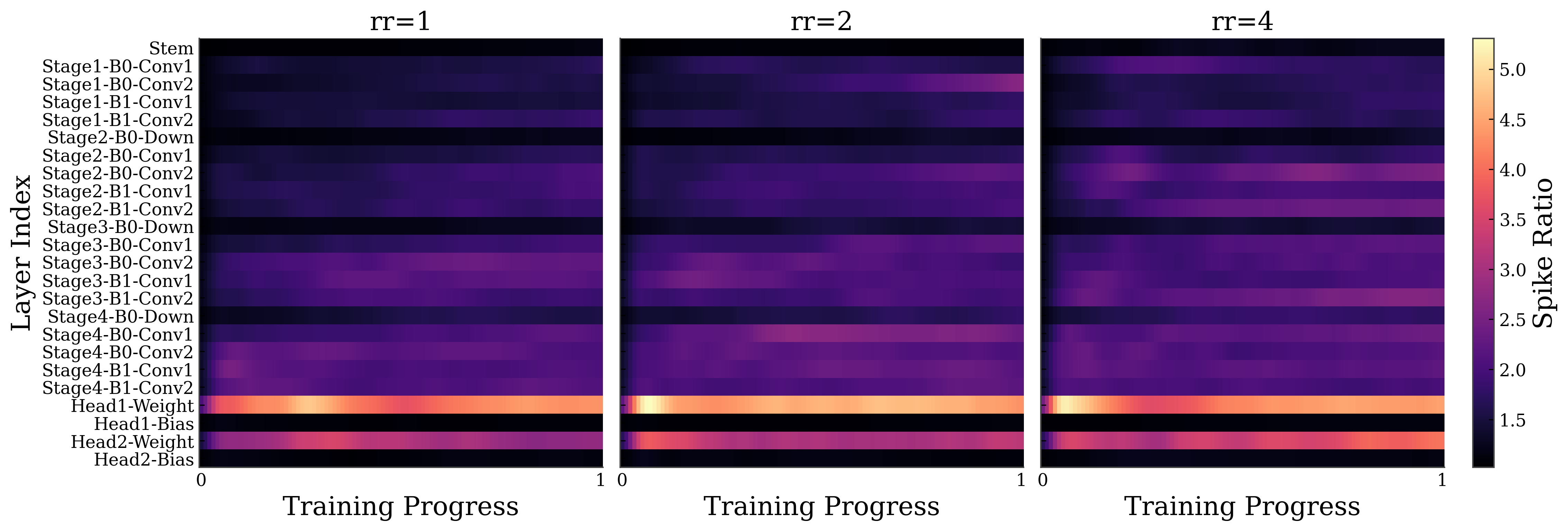}
\caption{Layer-wise spike ratio evolution during training under different replay ratios. Each row corresponds to a network layer, and color intensity indicates spike magnitude.}
\label{fig: ab_spikeratio}
\end{figure}
This section provides detailed experimental settings and implementation details for the spike ratio analysis presented in Fig.~\ref{fig: ab_spikeratio}(Fig.~2 of the main paper). The purpose of this experiment is to examine how replay ratio evolves over the course of learning.

To monitor parameter dynamics during training, we periodically compute the spike ratio.
For a network layer $\ell$ with parameter tensor $\theta_\ell$, the spike ratio is defined as
\begin{equation}
\text{SpikeRatio}_\ell =
\frac{\|\theta_\ell\|_{\infty}}
{\text{Quantile}_{0.99}(|\theta_\ell|)}.
\end{equation}

The experiment follows the standard Atari-100K evaluation protocol described in the main paper.
Training uses the same optimization settings as the main algorithm described in this paper.
Unless otherwise specified, all hyperparameters remain identical to the main experiments.
The replay ratio $rr$ controls the number of gradient updates per environment step.
To analyze the effect of data reuse, we conduct experiments with three different settings, $rr \in \{1,2,4\}$.

The value function is implemented as an ensemble quantile network.
Each ensemble member consists of a convolutional encoder followed by a linear projection head that outputs quantile estimates for all actions.

The encoder adopts a lightweight ResNet-style architecture.
Specifically, the input observation $x \in \mathbb{R}^{4\times84\times84}$ is processed by a sequence of convolutional layers and residual blocks to produce a latent feature vector.

\begin{table}[h]
\centering
\caption{Network layers monitored for spike ratio statistics in Figure~\ref{fig: ab_spikeratio}. The spike ratio is computed for each layer every 1000 environment steps to analyze how parameter imbalance evolves during training. The encoder's channel width is controlled by a scale factor $s$ (with $s=4$ in our experiments).}
\label{tab:spike_layers}
\begin{tabularx}{\linewidth}{l l
    >{\centering\arraybackslash}X
    >{\centering\arraybackslash}X}
\toprule
\textbf{Layer Group} &
\textbf{Layer Name} &
\textbf{Input Shape} &
\textbf{Weight Shape} \\
\midrule

Stem
& stem.0
& $4 \times 84 \times 84$
& $(8s) \times 4 \times 3 \times 3$ \\

\midrule

Stage1
& l1.0.conv1
& $(8s) \times 84 \times 84$
& $(8s) \times (8s) \times 3 \times 3$ \\

& l1.0.conv2
& $(8s) \times 84 \times 84$
& $(8s) \times (8s) \times 3 \times 3$ \\

& l1.1.conv1
& $(8s) \times 84 \times 84$
& $(8s) \times (8s) \times 3 \times 3$ \\

& l1.1.conv2
& $(8s) \times 84 \times 84$
& $(8s) \times (8s) \times 3 \times 3$ \\

\midrule

Stage2
& l2.0.conv1
& $(8s) \times 84 \times 84$
& $(16s) \times (8s) \times 3 \times 3$ \\

& l2.0.conv2
& $(16s) \times 42 \times 42$
& $(16s) \times (16s) \times 3 \times 3$ \\

& l2.1.conv1
& $(16s) \times 42 \times 42$
& $(16s) \times (16s) \times 3 \times 3$ \\

& l2.1.conv2
& $(16s) \times 42 \times 42$
& $(16s) \times (16s) \times 3 \times 3$ \\

\midrule

Stage3
& l3.0.conv1
& $(16s) \times 42 \times 42$
& $(32s) \times (16s) \times 3 \times 3$ \\

& l3.0.conv2
& $(32s) \times 21 \times 21$
& $(32s) \times (32s) \times 3 \times 3$ \\

& l3.1.conv1
& $(32s) \times 21 \times 21$
& $(32s) \times (32s) \times 3 \times 3$ \\

& l3.1.conv2
& $(32s) \times 21 \times 21$
& $(32s) \times (32s) \times 3 \times 3$ \\

\midrule

Stage4
& l4.0.conv1
& $(32s) \times 21 \times 21$
& $(64s) \times (32s) \times 3 \times 3$ \\

& l4.0.conv2
& $(64s) \times 11 \times 11$
& $(64s) \times (64s) \times 3 \times 3$ \\

& l4.1.conv1
& $(64s) \times 11 \times 11$
& $(64s) \times (64s) \times 3 \times 3$ \\

& l4.1.conv2
& $(64s) \times 11 \times 11$
& $(64s) \times (64s) \times 3 \times 3$ \\

\midrule

Projection Head
& head.1
& $(64s) \times 11 \times 11$
& $(128s) \times (64s \cdot 11 \cdot 11)$ \\

Value Head
& quantile head
& $128s$
& $(|\mathcal{A}|K) \times (128s)$ \\

\bottomrule
\end{tabularx}

\end{table}
Table~\ref{tab:spike_layers} lists all trainable layers for which spike statistics are monitored during training. Since the model contains multiple ensemble members with identical architectures, spike ratios are first computed independently for each ensemble network and then averaged to obtain the aggregated statistics used in Figure~\ref{fig: ab_spikeratio}.

\section{Bounded Bellman Update}
This bound is derived from the structure of the discounted return and reflects the maximum achievable value under the sampled reward scale.

Consider a trajectory generated by greedily following the current value function.
Starting from $(s_0,a_0)$ with $a_0 \in \arg\max_{a} Q(s_0,a)$, we iteratively select
\[
a_t \in \arg\max_{a} Q(s_t,a),
\]
and obtain a greedy trajectory
\[
\mathcal{T}=\{(s_0,a_0,r_0),(s_1,a_1,r_1),\ldots,(s_T,a_T,r_T)\},
\]
which terminates at time step $T$.

Along this greedy trajectory, the one-step greedy backup satisfies
\begin{equation}
\label{eq:greedy_backup_on_traj}
r_t + \gamma \max_{a'} Q(s_{t+1},a')
\;=\;
r_t + \gamma Q(s_{t+1},a_{t+1}),
\end{equation}
because $a_{t+1}\in\arg\max_{a}Q(s_{t+1},a)$ by construction.
Therefore, by recursively substituting the greedy action at each subsequent state, we obtain
\begin{align}
\label{eq:q_rollout_expand}
\begin{split}
Q(s_t,a_t)
&= r_t + \gamma \max_{a'} Q(s_{t+1},a') \\
&= r_t + \gamma Q(s_{t+1},a_{t+1}) \\
&= r_t + \gamma\Big(r_{t+1} + \gamma \max_{a'} Q(s_{t+2},a')\Big) \\
&= r_t + \gamma r_{t+1} + \gamma^2 Q(s_{t+2},a_{t+2}) \\
&\;\;\vdots \\
&= r_t + \gamma r_{t+1} + \cdots + \gamma^{T-t} Q(s_T,a_T).
\end{split}
\end{align}
If the episode terminates at $T$ with no future continuation (or equivalently $Q(s_T,a_T)=r_T$ under terminal dynamics), the recursion closes and yields
\begin{equation}
\label{eq:q_equals_return_on_greedy_traj}
Q(s_t,a_t)
=
\sum_{k=t}^{T}\gamma^{k-t} r_k
\;\triangleq\;
G_t.
\end{equation}
For a given greedy trajectory $\mathcal{T}$, we define the trajectory-level value bound as
\begin{equation}
B(\mathcal{T}) = \max_{0 \le t \le T} G_t.
\end{equation}
This greedy-rollout identity provides a natural way to use $G_t$ as a trajectory-wise bound for the numerical scale of $Q(s_t,a_t)$ under the current greedy policy.

\section{Atari Experiment Details}
\setcounter{subsection}{0}

This section provides additional implementation details for the Atari experiments reported in the main paper. The experiments follow the Atari-100K evaluation protocol with a total interaction budget of 100k environment steps.

\subsection{Environment Setup}

All experiments are conducted using the Atari Learning Environment through Gymnasium. Observations follow the standard preprocessing pipeline implemented by the Stable-Baselines3 Atari wrapper.

Specifically, raw frames are converted to grayscale, resized to $84 \times 84$, and stacked over four consecutive frames to form the input state representation. Frame skipping is set to $4$, which corresponds to $400k$ environment frames under the Atari-100K setting.

The environment wrapper additionally applies a random number of no-op actions at the beginning of each episode (up to 30 steps).
Episode termination follows the standard Atari convention, where life loss is treated as terminal during training.

Rewards are not clipped to $[-1,1]$ during environment interaction.
However, the sign of the reward is stored in the replay buffer for training updates.

\subsection{Replay Buffer and Sampling}

Experience replay uses a prioritized replay buffer with a capacity of $100,000$ transitions.

Transitions are sampled according to prioritized sampling with exponent $\alpha = 0.6$. Importance sampling weights are applied during optimization using exponent $\beta = 0.4$.

For each sampled transition, observations are normalized to $[0,1]$ by dividing pixel values by $255$.

\subsection{Network Architecture}
As Table~\ref{tab:spike_layers}, the value function is implemented as an ensemble quantile network.
Each ensemble member consists of a convolutional encoder followed by a linear projection head.

The encoder adopts a lightweight ResNet-style architecture.
The input observation has shape $(4,84,84)$ and is processed through a sequence of convolutional residual blocks with progressive downsampling:
\[
84 \times 84 \rightarrow
84 \times 84 \rightarrow
42 \times 42 \rightarrow
21 \times 21 \rightarrow
11 \times 11.
\]
The number of channels is controlled by a scale factor $s=4$.
After the convolutional encoder, features are flattened and projected to a fully connected layer of size $512s$.

The final layer outputs $K$ quantile values for each action.
With $N_e$ ensemble members and $K$ quantile atoms, the network produces a tensor of shape
\[
(B, N_e, K, |\mathcal{A}|),
\]
where $B$ denotes the batch size.

\subsection{Training Procedure}

The agent is trained for $100,000$ environment steps.
During training, the agent interacts with the environment using an $\epsilon$-greedy policy.

The exploration parameter $\epsilon$ is linearly annealed from $1.0$ to $0.01$ over the first $2000$ steps.

Learning begins after $2000$ transitions have been collected.
At each environment step, $rr$ gradient updates are performed, where $rr$ denotes the replay ratio.

Optimization uses the Adam optimizer with learning rate $1\times10^{-4}$.

Target networks are updated using Polyak averaging with coefficient $\tau = 0.005$.

Gradients are clipped with a maximum norm of $10$.

\subsection{Distributional Value Learning}

The value function is trained using quantile regression with the quantile Huber loss. The number of quantile atoms is $K=51$.

Quantile fractions follow the standard QR-DQN formulation
\[
\tau_i = \frac{i+0.5}{K}, \quad i=0,\dots,K-1.
\]
For each update, the target distribution is constructed using the target network and the greedy action selected according to the mean value across quantiles.

\subsection{Ensemble Learning}

The proposed method employs an ensemble of $N_e=16$ value networks.
Each ensemble member independently estimates quantile value distributions for all actions.

During action selection, quantile values are averaged across both the quantile dimension and the ensemble dimension to produce the final action-value estimates.

\subsection{Parameter Reset Mechanism}

To mitigate excessive parameter specialization, the algorithm monitors the spike ratio of each layer during training.

For a parameter tensor $\theta$, the spike ratio is defined as

\[
\text{SpikeRatio} =
\frac{\max |\theta|}
{\text{Quantile}_{0.99}(|\theta|)}.
\]

If the spike ratio of a layer exceeds the threshold, the parameters of that layer are reinitialized.

The optimizer states associated with the reset parameters are also cleared, and the target network is synchronized with the updated parameters.

\subsection{Hardware and Implementation}

All experiments were conducted on a single NVIDIA RTX~5090 GPU.
The implementation is written in PyTorch and runs on a Linux environment with CUDA acceleration.
All reported results are obtained using the same hardware configuration.

\subsection{Hyperparameters}
The hyperparameters used in the Atari experiments are summarized in Table~\ref{tab: atari_hyper}.

\begin{table}[h]
\centering
\caption{Hyperparameter configuration used for Atari-100K experiments.}
\label{tab: atari_hyper}
\begin{tabularx}{\linewidth}{l
>{\centering\arraybackslash}X
}
\toprule
\multicolumn{2}{c}{Training Configuration (Atari-100K)} \\
\midrule
Total environment steps          & $100{,}000$ \\
Discount factor $\gamma$         & $0.99$ \\
Learning rate                    & $1\times10^{-4}$ (Adam) \\
Batch size                       & $32$ \\
Replay buffer size               & $100{,}000$ \\
Start learning after steps       & $2{,}000$ \\
Training frequency               & $1$ update per step \\
Soft target update rate $\tau$   & $0.005$ \\
Replay ratio                     & $4$ \\
Gradient clipping                & $10.0$ \\
Parameter reset threshold        & $6.0$ \\
\midrule
\multicolumn{2}{c}{Exploration} \\
\midrule
Initial $\epsilon$               & $1.0$ \\
Final $\epsilon$                 & $0.01$ \\
$\epsilon$ decay steps           & $2{,}001$ \\
\midrule
\multicolumn{2}{c}{Replay Buffer} \\
\midrule
Prioritization exponent $\alpha$ & $0.6$ \\
Importance sampling exponent $\beta$ & $0.4$ \\
Priority epsilon                 & $1\times10^{-6}$ \\
\midrule
\multicolumn{2}{c}{Distributional / Ensemble Settings} \\
\midrule
Number of quantiles              & $51$ \\
Number of ensemble heads         & $16$ \\
Quantile Huber parameter $\kappa$& $1$ \\
\midrule
\multicolumn{2}{c}{Observation Processing} \\
\midrule
Frame stack                      & $4$ \\
Frame skip                       & $4$ \\
Observation resolution           & $84\times84$ \\
\bottomrule
\end{tabularx}
\end{table}

\subsection{Experimental Results}

Detailed performance comparisons on the Atari benchmark
are provided in Table~\ref{tab: atari_full}.
\begin{sidewaystable}[p]
\centering
\small
\caption{Atari-100K performance comparison across environments. The best score in each row is highlighted in bold.}
\label{tab: atari_full}
\setlength{\tabcolsep}{3pt}
% \begin{tabularx}{\textwidth}{
% l
% >{\centering\arraybackslash}X
% >{\centering\arraybackslash}X
% >{\centering\arraybackslash}X
% >{\centering\arraybackslash}X
% >{\centering\arraybackslash}X
% >{\centering\arraybackslash}X
% >{\centering\arraybackslash}X
% >{\centering\arraybackslash}X
% >{\centering\arraybackslash}X
% >{\centering\arraybackslash}X
% >{\centering\arraybackslash}X
% >{\centering\arraybackslash}X
% >{\centering\arraybackslash}X
% >{\centering\arraybackslash}X
% }
% \begin{tabularx}{\textheight}{p{1cm} *{14}{>{\centering\arraybackslash}X}}
\begin{tabularx}{\textheight}{p{2.5cm} *{14}{c}}
\toprule
\textbf{Env} &
\textbf{Random} &
\textbf{Human} &
\textbf{DER} &
\textbf{DrQ} &
\textbf{IRIS} &
\textbf{REM} &
\textbf{STORM} &
\textbf{DreamerV3} &
\textbf{DIAMOND} &
\textbf{DART} &
\textbf{SGF} &
\textbf{Drama} &
\textbf{BBF} &
\textbf{Ours} \\
\midrule
Alien          & 227.8 & 7127.7 & 802.3 & 865.2 & 420.0 & 607.2 & 983.6 & 959.0 & 744.1 & 962.0 & 518.8 & 820 & 1173.2 & \textbf{1340.0} \\
Amidar         & 5.8 & 1719.5 & 125.9 & 137.8 & 143.0 & 95.3 & 204.8 & 139.0 & 225.8 & 125.7 & 62.7 & 131 & 244.6 & \textbf{293.6} \\
Assault        & 222.4 & 742.0 & 561.5 & 579.6 & 1524.4 & 1764.2 & 801.0 & 706.0 & 1526.4 & 1316.0 & 850.1 & 539 & 2098.5 & \textbf{2348.3} \\
Asterix        & 210.0 & 8503.3 & 535.4 & 763.6 & 853.6 & 1637.5 & 1028.0 & 932.0 & 3698.5 & 956.2 & 802.5 & 1632 & \textbf{3946.1} & 2430.0 \\
BankHeist      & 14.2 & 753.1 & 185.5 & 232.9 & 53.1 & 19.2 & 641.2 & 649.0 & 19.7 & 629.7 & 58.7 & 137 & 732.9 & \textbf{772.0} \\
BattleZone     & 2360.0 & 37187.5 & 8977.0 & 10165.3 & 13074.0 & 11826.0 & 13540.0 & 12250.0 & 4702.0 & 15325.0 & 3747.0 & 10860 & \textbf{24459.8} & 18600.0 \\
Boxing         & 0.1 & 12.1 & -0.3 & 9.0 & 70.1 & \textbf{87.5} & 79.7 & 78.0 & 86.9 & 83.0 & 83.4 & 78 & 85.8 & 25.1 \\
Breakout       & 1.7 & 30.5 & 9.2 & 19.8 & 83.7 & 90.7 & 15.9 & 31.0 & 132.5 & 41.9 & 50.7 & 7 & \textbf{370.6} & 233.0 \\
ChopperCommand & 811.0 & 7387.8 & 925.9 & 844.6 & 1565.0 & 2561.2 & 1888.0 & 420.0 & 1369.8 & 1263.8 & 1775.4 & 1642 & \textbf{7549.3} & 2740.0 \\
CrazyClimber   & 10780.5 & 35829.4 & 34508.6 & 21539.0 & 59324.2 & 76547.6 & 66776.0 & \textbf{97190.0} & 99167.8 & 34070.6 & 15751.3 & 83931 & 58431.8 & 84700.0 \\
DemonAttack    & 152.1 & 1971.0 & 627.6 & 1321.5 & 2034.4 & 5738.6 & 164.6 & 303.0 & 288.1 & 2452.3 & 2809.5 & 201 & \textbf{13341.4} & 7192.5 \\
Freeway        & 0.0 & 29.6 & 20.9 & 20.3 & 31.1 & 32.3 & \textbf{33.5} & 0.0 & 33.3 & 32.2 & 11.9 & 15 & 25.5 & 33.3 \\
Frostbite      & 65.2 & 4334.7 & 871.0 & 1014.2 & 259.1 & 240.5 & 1316.0 & 909.0 & 274.1 & 346.8 & 265.6 & 785 & 2384.8 & \textbf{2886.0} \\
Gopher         & 257.6 & 2412.5 & 467.0 & 621.6 & 2236.1 & 5452.4 & \textbf{8239.6} & 3730.0 & 5897.9 & 1980.5 & 416.4 & 2757 & 1331.2 & 2346.0 \\
Hero           & 1027.0 & 30826.4 & 6226.0 & 4167.9 & 7037.4 & 6484.8 & 11044.3 & 11161.0 & 5621.8 & 4927.0 & 1522.9 & 7946 & 7818.6 & \textbf{13436.0} \\
Jamesbond      & 29.0 & 302.8 & 275.7 & 349.1 & 462.7 & 391.2 & 509.0 & 445.0 & 427.4 & 353.1 & 280.9 & 372 & \textbf{1129.6} & 510.0 \\
Kangaroo       & 52.0 & 3035.0 & 581.7 & 1088.4 & 838.2 & 467.6 & 4208.0 & 4098.0 & 5382.2 & 2380.0 & 271.2 & 1384 & 6614.7 & \textbf{7520.0} \\
Krull          & 1598.0 & 2665.5 & 3256.9 & 4402.1 & 6616.4 & 4017.7 & 8412.6 & 7782.0 & 8610.1 & 7658.3 & 7813.7 & \textbf{9693} & 8223.4 & 7851.0 \\
KungFuMaster   & 258.5 & 22736.3 & 6580.1 & 11467.4 & 21759.8 & 25172.2 & \textbf{26182.0} & 21420.0 & 18713.6 & 23744.3 & 20169.8 & 23920 & 18991.7 & 24190.0 \\
MsPacman       & 307.3 & 6951.6 & 1187.4 & 1218.1 & 999.1 & 962.5 & \textbf{2673.5} & 1327.0 & 1958.2 & 1132.7 & 1356.8 & 2270 & 2008.3 & 1986.0 \\
Pong           & -20.7 & 14.6 & -9.7 & -9.1 & 14.6 & 18.0 & 11.3 & 18.0 & 20.4 & 17.2 & 12.6 & 15 & 16.7 & \textbf{21.0} \\
PrivateEye     & 24.9 & 69571.3 & 72.8 & 3.5 & 100.0 & 99.6 & \textbf{7781.0} & 882.0 & 114.3 & 765.7 & 405.5 & 90 & 40.5 & 100.0 \\
Qbert          & 163.9 & 13455.0 & 1773.5 & 1810.7 & 745.7 & 743.0 & 4522.5 & 3405.0 & 4499.3 & 750.9 & 685.0 & 796 & 4447.1 & \textbf{4985.0} \\
RoadRunner     & 11.5 & 7845.0 & 11843.4 & 11211.4 & 9614.6 & 14060.2 & 17564.0 & 15565.0 & 20673.2 & 7772.5 & 8164.2 & 14020 & \textbf{33426.8} & 32760.0 \\
Seaquest       & 68.4 & 42054.7 & 304.6 & 352.3 & 661.3 & 1036.7 & 525.2 & 618.0 & 551.2 & 895.8 & 476.8 & 497 & \textbf{1232.5} & 1198.0 \\
UpNDown        & 533.4 & 11693.2 & 3075.0 & 4324.5 & 3546.2 & 3757.6 & 7985.0 & 9234.0 & 3856.3 & 3954.5 & 7745.0 & 7387 & 12101.7 & \textbf{33994.0} \\
\midrule
\textbf{\# > Human} & 0 & 0 & 2 & 3 & 9 & 12 & 10 & 9 & 11 & 9 & 6 & 8 & 12 & 14 \\
\textbf{\# Best}    & 0 & 0 & 0 & 0 & 0 & 1 & 5  & 1 & 0 & 0 & 0 & 1 & 8  & 10  \\
\textbf{Mean HNS}   & 0.000 & 1.000 & 0.350 & 0.465 & 1.046 & 1.222 & 1.266 & 1.120 & 1.459 & 1.022 & 0.884 & 1.049 & 2.247 & 1.799 \\
\textbf{Median HNS} & 0.000 & 1.000 & 0.189 & 0.313 & 0.289 & 0.280 & 0.580 & 0.466 & 0.373 & 0.790 & 0.152 & 0.270 & 0.917 & 1.045 \\
\textbf{IQM HNS}    & 0.000 & 1.000 & 0.183 & 0.280 & 0.501 & 0.673 & 0.636 & 0.490 & 0.641 & 0.575 & 0.287 & 0.367 & 1.045 & 1.070 \\
\bottomrule
\end{tabularx}
\end{sidewaystable}

\section{Procgen Experiment Details}
\setcounter{subsection}{0}

This section provides additional implementation details for the Procgen experiments reported in the main paper.
All experiments follow the standard Procgen generalization protocol.

\subsection{Environment Setup}

We evaluate our method on the Procgen benchmark under the generalization mode.
In this setting, the agent is trained on a fixed set of procedurally generated levels and evaluated on unseen levels.

Specifically, training environments use the first 200 levels of each game (\texttt{start\_level = 0}, \texttt{num\_levels = 200}), while evaluation is performed on disjoint levels starting from level 200 (\texttt{start\_level = 200}, \texttt{num\_levels = 0}), which corresponds to an infinite stream of unseen levels.

All environments are run in \texttt{easy} difficulty mode.
Observations are RGB images with resolution $64\times64$.

Following common practice in prior work, we use a vectorized environment with 128 parallel instances during training to improve data throughput.

\subsection{Observation Processing}

Unlike Atari, Procgen observations are already provided as RGB images with fixed resolution.
Therefore no grayscale conversion or resizing is required.

Observations are represented as tensors of shape $(3, 64, 64)$ corresponding to RGB channels.
Pixel values are normalized to $[0,1]$ before being fed into the neural network.

Frame stacking is not used in our experiments since Procgen environments are fully observable.

\subsection{Network Architecture}

The value function is implemented as the same ensemble quantile network used in the Atari experiments, but with an encoder adapted to the $64\times64$ Procgen observations.

The encoder follows a lightweight residual architecture.
Starting from a $64\times64$ input image, the spatial resolution is progressively reduced through strided residual blocks:
\[
64 \times 64
\rightarrow
32 \times 32
\rightarrow
16 \times 16
\rightarrow
8 \times 8
\rightarrow
4 \times 4.
\]
The resulting feature map is flattened and projected to a 512-dimensional feature vector.
The final layer outputs $K$ quantile values for each action.

With $N_e$ ensemble members and $K$ quantiles, the network produces a tensor of shape
\[
(B, N_e, K, |\mathcal{A}|),
\]
where $B$ denotes the batch size.

\subsection{Training Procedure}

Agents are trained for a total of 25 million environment steps.

Exploration follows an $\epsilon$-greedy strategy. The exploration rate is linearly annealed from $1.0$ to $0.05$ over the first $5$ million environment steps.

Training begins after an initial replay buffer warm-up period.
During training, the agent performs gradient updates using mini-batches sampled from the replay buffer.

The optimization objective is the quantile regression loss.
The Adam optimizer is used for parameter updates.

\subsection{Replay}

The replay buffer stores up to $100,000$ transitions.
Mini-batches of size 256 are sampled uniformly during training.

\subsection{Ensemble Distributional Value Learning}

Our method employs an ensemble of $N_e = 16$ value networks.
Each ensemble member independently estimates the quantile value distribution for each action.

During action selection, the predicted quantiles are averaged across both the ensemble dimension
and the quantile dimension to produce the final action-value estimates.

\subsection{Hardware and Implementation}

All Procgen experiments are implemented in PyTorch and executed on a single NVIDIA RTX 5090 GPU.
The implementation uses vectorized environments with CUDA acceleration to improve training throughput.

The hyperparameters used for the Procgen experiments are summarized in Table~\ref{tab:procgen_hyper}.
\begin{table}[h]
\centering
\caption{Hyperparameters for Procgen (generalization mode, 200 training levels).}
\label{tab:procgen_hyper}
\begin{tabularx}{\linewidth}{l
>{\centering\arraybackslash}X
}
\toprule
Total environment steps & 25M \\
Training levels & 200 \\
Evaluation levels & Unseen procedural levels \\
Parallel environments & 128 \\
Batch size & 256 \\
Discount factor $\gamma$ & 0.99 \\
Optimizer & Adam \\
Learning rate & 1e-4 \\
Soft target update rate $\tau$   & $0.005$ \\
Ensemble size $N_e$ & 16 \\
Quantile number $K$ & 51 \\
Initial exploration $\epsilon$ & $1.0$ \\
Final exploration $\epsilon$& $0.05$ \\
$\epsilon$ decay steps & $5M$ \\
\bottomrule
\end{tabularx}
\end{table}

\subsection{Detailed Per-Environment Results}

To provide a more comprehensive view of performance across different tasks, we report the raw test scores for each Procgen environment in Table~\ref{tab:procgen_full} under the generalization protocol.
\begin{table}[h]
\centering
\caption{Per-environment raw test performance on Procgen (easy, 200 training levels).}
\label{tab:procgen_full}
\begin{tabularx}{\linewidth}{l
>{\centering\arraybackslash}X
>{\centering\arraybackslash}X
}
\toprule
Environment & PPO (Impala CNN) & Ours \\
\midrule
BigFish   & 3.0  & 18.1 \\
BossFight & 8.3  & 8.1 \\
CaveFlyer & 5.6  & 7.0 \\
Chaser    & 5.5  & 1.9 \\
Climber   & 6.1  & 7.5 \\
CoinRun   & 8.8  & 6.0 \\
Dodgeball & 2.2  & 16.0 \\
FruitBot  & 27.0 & 25.7 \\
Heist     & 2.2  & 1.0 \\
Jumper    & 5.8  & 6.0 \\
Leaper    & 4.4  & 6.0 \\
Maze      & 5.7  & 2.0 \\
Miner     & 9.0  & 6.1 \\
Ninja     & 5.8  & 6.0 \\
Plunder   & 5.4  & 5.5 \\
StarPilot & 25.0 & 54.9 \\
\bottomrule
\end{tabularx}
\end{table}

\begin{figure}[h]
\centering
\includegraphics[width=\linewidth]{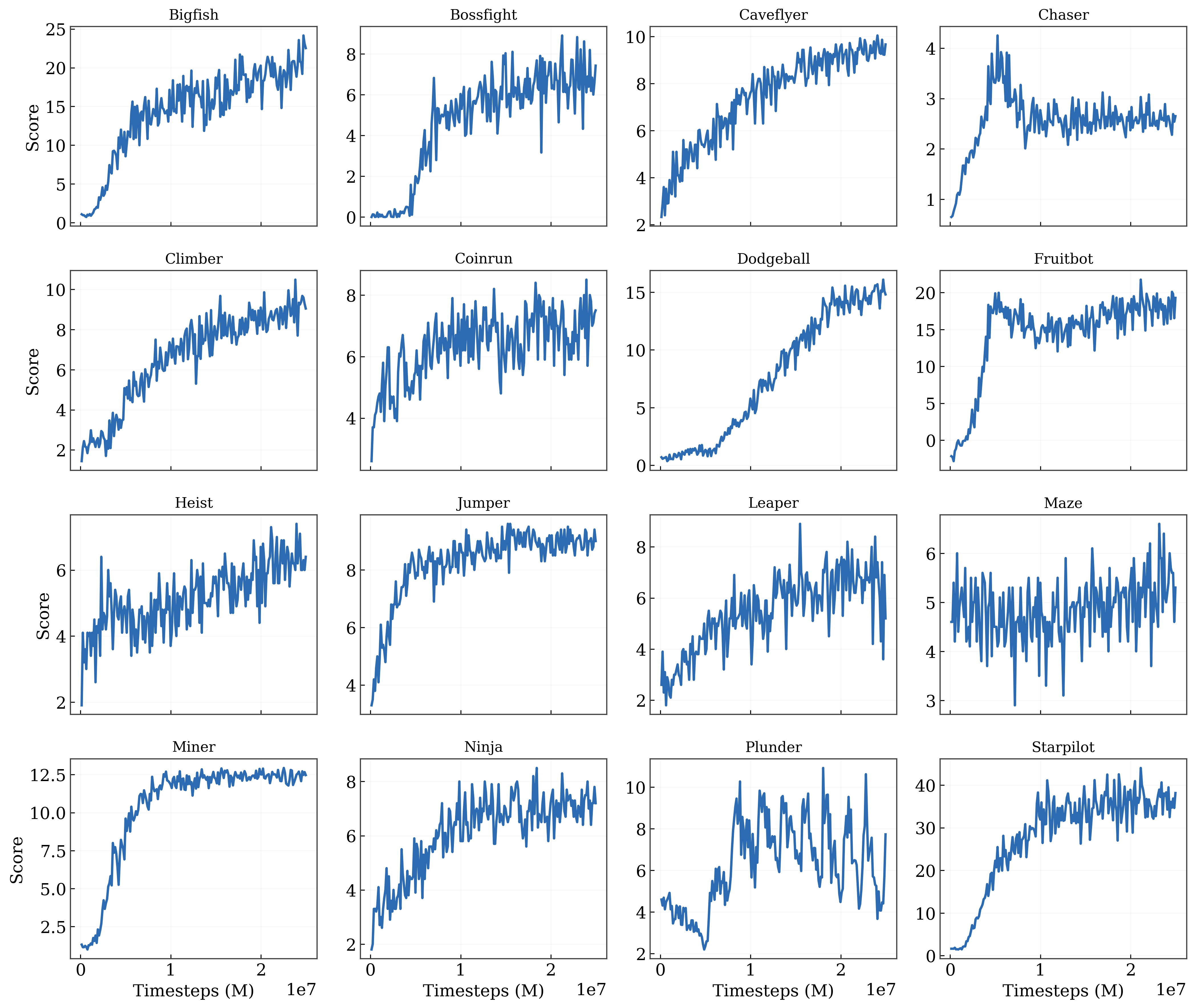}
\caption{Per-environment train performance during training on Procgen.}
\label{fig:procgen_curves}
\end{figure}

In addition to the final evaluation scores, we also visualize the training dynamics for each environment.
Figure~\ref{fig:procgen_curves} shows the training performance curves throughout the entire learning process.
Each subplot corresponds to a different Procgen environment, and the horizontal axis represents the number of environment interaction steps.

\end{document}